\documentclass[10pt,twocolumn,letterpaper]{article}

\usepackage{cvpr}              
\definecolor{cvprblue}{rgb}{0.21,0.49,0.74}
\usepackage[pagebackref,breaklinks,colorlinks,allcolors=cvprblue]{hyperref}

\usepackage{multirow}

\def\paperID{*****} 
\def\confName{CVPR}
\def\confYear{2026}

\title{\includegraphics[width=0.04\linewidth]{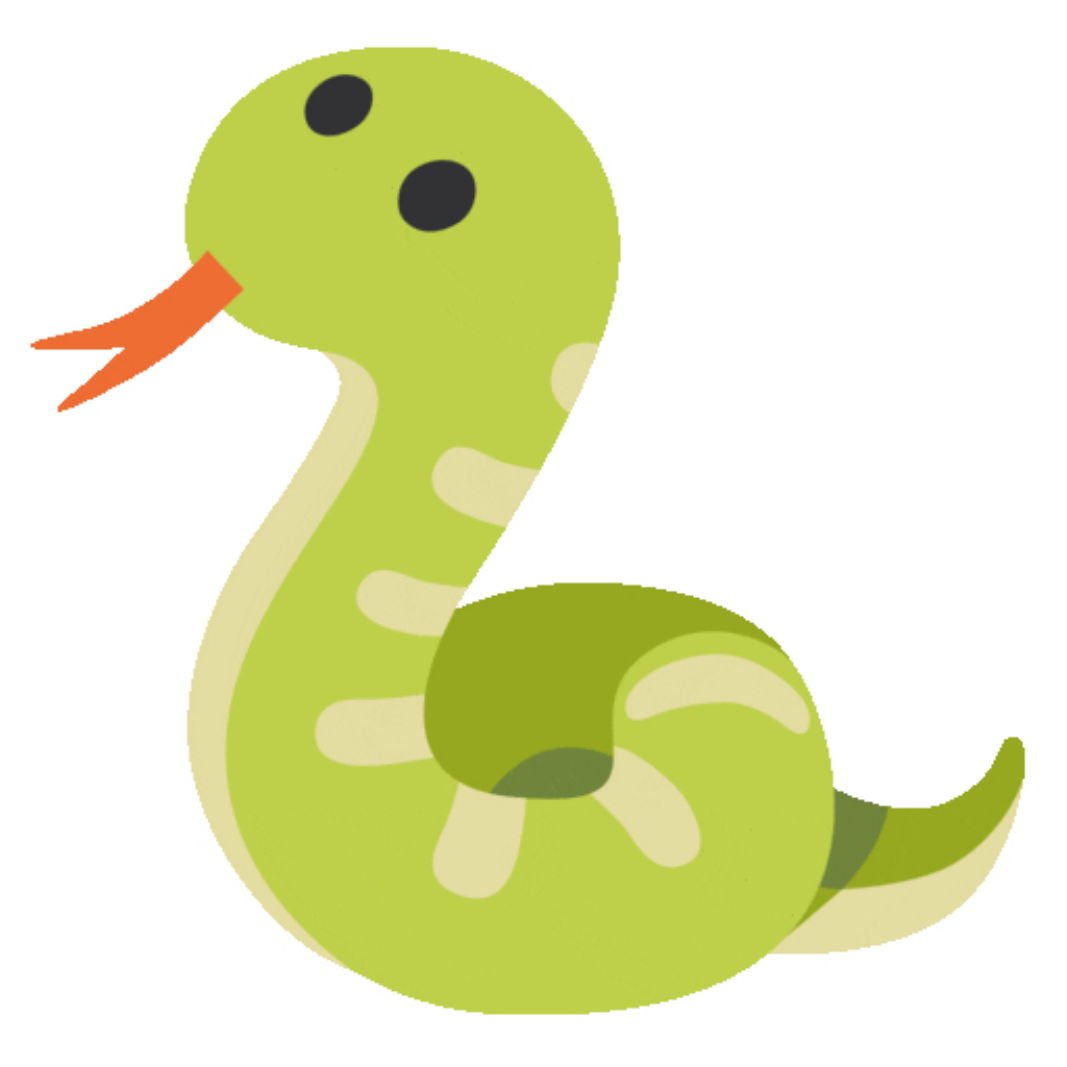} DA-Mamba: Learning Domain-Aware State Space Model for Global-Local Alignment in Domain Adaptive Object Detection}

\author{
    Haochen~Li\textsuperscript{1, 4}\quad Rui~Zhang\textsuperscript{2}\thanks{Corresponding author.}\quad Hantao~Yao\textsuperscript{3}\quad
    Xin~Zhang\textsuperscript{2}\quad \\
    Yifan~Hao\textsuperscript{2}\quad  
    Shaohui~Peng\textsuperscript{1}\quad 
    Yongwei~Zhao\textsuperscript{2}\quad 
    Ling~Li\textsuperscript{1, 4}\footnote[1]{Corresponding author.}\\
    \textsuperscript{1}Intelligent Software Research Center, Institute of Software, CAS\\
    \textsuperscript{2}State Key Lab of Processors, Institute of Computing Technology, CAS\\
    \textsuperscript{3}
School of Information Science and Technology, University of Science and Technology of China\\
    \textsuperscript{4} University of Chinese Academy of Sciences \\
    {\tt\small  \{haochen2021, pengshaohui, liling\}@iscas.ac.cn} \\
    {\tt\small  \{zhangrui, zhangxin, haoyifan, zhaoyongwei\}@ict.ac.cn, yaohantao@ustc.edu.cn}\\
    \vspace{-30pt}
}

\begin{document}

\maketitle

\begin{abstract}
Domain Adaptive Object Detection (DAOD) aims to transfer detectors from a labeled source domain to an unlabeled target domain.
Existing DAOD methods employ multi-granularity feature alignment to learn domain-invariant representations.
However, the local connectivity of their CNN-based backbone and detection head restricts alignment to local regions, failing to extract global domain-invariant features.
Although transformer-based DAOD methods capture global dependencies via attention mechanisms, their quadratic computational cost hinders practical deployment. 
To solve this, we propose DA-Mamba, a hybrid CNN-State Space Models (SSMs)  architecture that combines the efficiency of CNNs with the linear-time long-range modeling capability of State Space Models (SSMs) to capture both global and local domain-invariant features.
Specifically, we introduce two novel modules: Image-Aware SSM (IA-SSM) and Object-Aware SSM (OA-SSM).
IA-SSM is integrated into the backbone to enhance global domain awareness, enabling image-level global and local alignment.
OA-SSM is inserted into the detection head to model spatial and semantic dependencies among objects, enhancing instance-level alignment.
Comprehensive experiments demonstrate that the proposed method can efficiently improve the cross-domain performance of the object detector.
\end{abstract}
\vspace{-20pt}

\section{Introduction}
\label{sec:intro}

\begin{figure*}[t]
\centering
\includegraphics[width=1.0\textwidth]{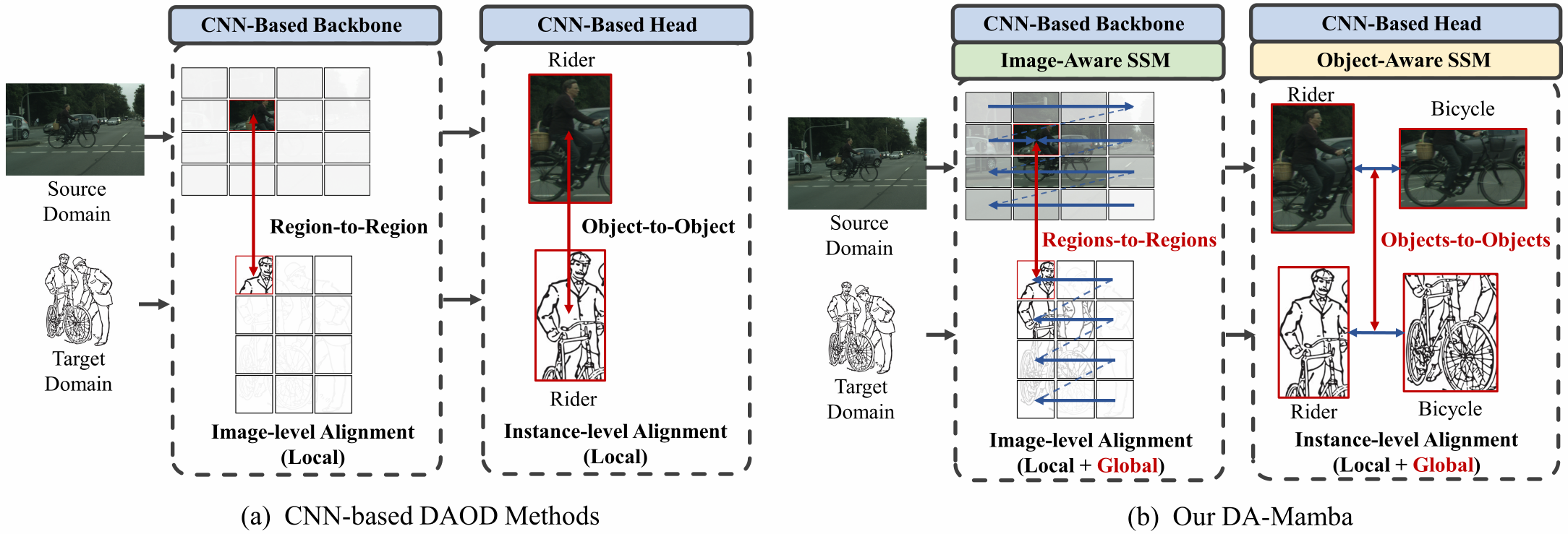} 
\vspace{-18pt}
\caption{
(a) Existing DAOD methods focus on aligning features at image and instance-level to extract domain-invariant features.
However, they only achieve local alignment due to the locality of CNN-based backbone and head, resulting in insufficient alignment across regions.
(b) Our DA-Mamba utilizes SSM's long-range perception capability to extract global domain-invariant features,
introducing the Image-Aware SSM module to the backbone to supplement global domain attributes, and the Object-Aware SSM module to the detection head to model spatial and semantic dependencies between objects, achieving fine-grained alignment at image and instance-level.}
\label{fig1}
\vspace{-15pt}
\end{figure*}

Traditional object detection algorithms achieve remarkable performance in both accuracy and speed, especially empowered by Convolutional Neural Networks (CNNs).
However, the inherent domain shift between task scenarios, such as object styles and scales, substantially degrades model generalization when transferring across domains.
To mitigate this issue, Domain Adaptive Object Detection (DAOD) leverages labeled data from a source domain to improve the detection performance in an unlabeled target domain.

A key idea behind DAOD is aligning features from both domains to accurately extract knowledge that can be shared between domains, \ie, domain-invariant features.
To achieve this, ~\cite{DA-Faster, Strong-weak, REACT} propose aligning the visual features extracted by the backbone, enabling the detector to focus on domain-invariant features at the image-level.
Furthermore, ~\cite{sigma++,CAT,CSDA} provide more fine-grained alignment from the perspective of categories and instances.
Recently, with the help of Visual-Language Models (VLMs)~\cite{CLIP, Regionclip, YOLO-W} to ground visual and textual modalities, VLM-based DAOD methods~\cite{DA-Pro, DA-Ada} explore aligning the domain semantics by tuning domain-adaptive prompts and adapters. 

Despite the success of multi-granularity alignment strategies in reducing feature discrepancies across domains, the local connectivity of convolutions restricts alignment to local regions, thus hindering global domain alignment.
As illustrated in Fig~\ref{fig1}(a), the CNN-based backbone captures only local features within the sliding window, failing to capture global visual attributes such as the relationships between different regions.
Similarly, the detection head fails to model global domain attributes, reflected in the spatial and semantic dependencies among all instances.
Spatially, objects often exhibit stable co-occurrence patterns—for example, a rider typically appears with a bicycle, while a table rarely co-occurs with a car—and semantically, instances such as cat and dog share strong conceptual relations as animals, while car and truck belong to the same category of vehicles.
Consequently, the ignorance of long-range dependencies across entire image causes existing region-to-region and object-to-object alignments to remain local, thereby limiting the learning of global domain-invariant features.

To address the locality limitation, ~\cite{DS, MTM, DATR} introduce Visual Transformer~\cite{Transformer, ViT} into DAOD, leveraging attention mechanism to capture global dependencies.
However, the quadratic complexity of self-attention leads to substantial computational and memory overhead~\cite{visionMamba}.
Although some convolution-based attention variants attempt to approximate global interactions more efficiently, they either rely on global parameter sharing(\eg a shared query), which ignores domain bias and hinders sufficient global adaptation, or still exhibit quadratic complexity in disguise.
Motivated by the success of Mamba~\cite{VMamba, U-Mamba} in modeling long-range dependencies with linear time complexity, it is promising to combine the State Space Models (SSMs)~\cite{Mamba} with CNN-based detectors to extract global domain information while maintaining high efficiency.

In this paper, we propose DA-Mamba for DAOD, innovatively integrating SSMs into CNN-based detectors to capture both local and global domain-invariant features.
Specifically, we design the Image-Aware SSM (IA-SSM) and Object-Aware SSM (OA-SSM) modules to perform local and global domain alignment at the image and instance-level, respectively.
As shown in Figure~\ref{fig1}, IA-SSM is inserted into the backbone to enhance long-range perception, enabling fine-grained alignment at the image level by simultaneously extracting local and global domain-invariant features.
The OA-SSM is integrated into the head to explore spatial dependencies among instances while learning semantic dependencies of categories in different domains, promoting instance-level alignment.
Empowered by the novel CNN-SSM hybrid architecture, DA-Mamba maintains high computational efficiency while supplementing the long-distance perception capability, showing superior robustness to various domain shifts.

Evaluations on four mainstream benchmarks highlight the effectiveness of the proposed DA-MAmba.
In particular, it achieves $58.1\%$ and $52.5\%$ mAP on Cityscapes$\rightarrow$Foggy and Pascal$\rightarrow$Clipart, outperforming the state-of-the-art by $2.2\%$ and $3.4\%$ with only $53.1\%$ of the FLOPs of Transformer-based DAOD methods.

\section{Related Work}
\label{sec:related}

\noindent\textbf{Domain Adaptive Object Detection (DAOD)} aims to leverage labeled data from a source domain to improve detection performance in an unlabeled target domain.
Existing work can be divided into three main categories: feature alignment, semi-supervised learning and VLM-based domain alignment.
As the mainstream paradigm, feature alignment involves aligning the source and target domains in feature space to learn domain-invariant representations.
Early methods such as DA-Faster~\cite{DA-Faster} employ a domain discriminator with a novel Gradient Reverse Layer(GRL) to adversarially align feature distributions at image-level and instance-level~\cite{RPN, Strong-weak}, while subsequent works~\cite{MEGA-CDA, SIGMA, CAT, REACT} further explore category-level alignment to preserve category semantic information across domains.
In parallel, semi-supervised learning reduces domain bias by generating pseudo labels~\cite{MT, AT, PT, HT} or using style transfer~\cite{FSAC, TDD, UMT} for unbiased data generation.
Recently, VLM-based methods~\cite{DA-Pro, DA-Ada} leverage the semantic priors of large-scale vision-language models to align the semantic commonalities between domains and supplement the specific information of the target domain.

Despite their success, these CNN-based approaches are inherently limited by local receptive fields, aligning only local features while neglecting global domain dependencies.
To mitigate this, ~\cite{DS, MTM, DATR} incorporate Visual Transformer~\cite{Transformer, ViT} with feature alignment methods, utilizing global attention mechanisms for long-range dependencies modeling.
However, the quadratic complexity and large parameter count of Transformers make them computationally expensive.
Although some convolution-based attention variants~\cite{GCNet,HAM,LAGConv} attempt to approximate global interactions more efficiently, they either rely on global parameter sharing (\eg, a shared query), which ignores domain bias and fails to discriminate inter-domain difference, or still incur quadratic complexity despite their linear appearance.

\noindent\textbf{State Space Model (SSMs)}
State space models are linear time-invariant systems that model dynamic input signal into output response, which can be traced back to the Kalman Filter~\cite{kalman_filter}.
Inspired by its linear sequence attention mechanism, ~\cite{flashattention, mega} explores it as an efficient and low-parameter alternative to Transformer~\cite{Transformer}.
~\cite{Hippo} comprehensively discusses the capability of SSMs in handling long-range dependencies, and further propose a practical Structured State Space Sequence (S4)~\cite{S4} model.
Following S4, ~\cite{S4_1d} introduce SSM to vision task like video classification,  exploring its potential in image sequence learning.
Motivated by Gated Recurrent Neural Network~\cite{GRU}, ~\cite{H3} propose H3 block, which forms the basis of the mainstream SSM network.
Recently,~\cite{Mamba} combined H3 block and Gated MLP to propose a concise and efficient linear attention module, Mamba.
Since its appearance, a series of Mamba models have been applied to various scenarios, such as image classification~\cite{VMamba, visionMamba} and medical segmentation~\cite{U-Mamba}.
In this work, we explore marrying the success of Mamba in modeling long-range dependencies to the DAOD task, injecting global domain information into the CNN-based detector without introducing quadratic complexity.

\section{Method}
\label{sec:method}

\begin{figure*}[t]
\centering
\includegraphics[width=0.9\textwidth]{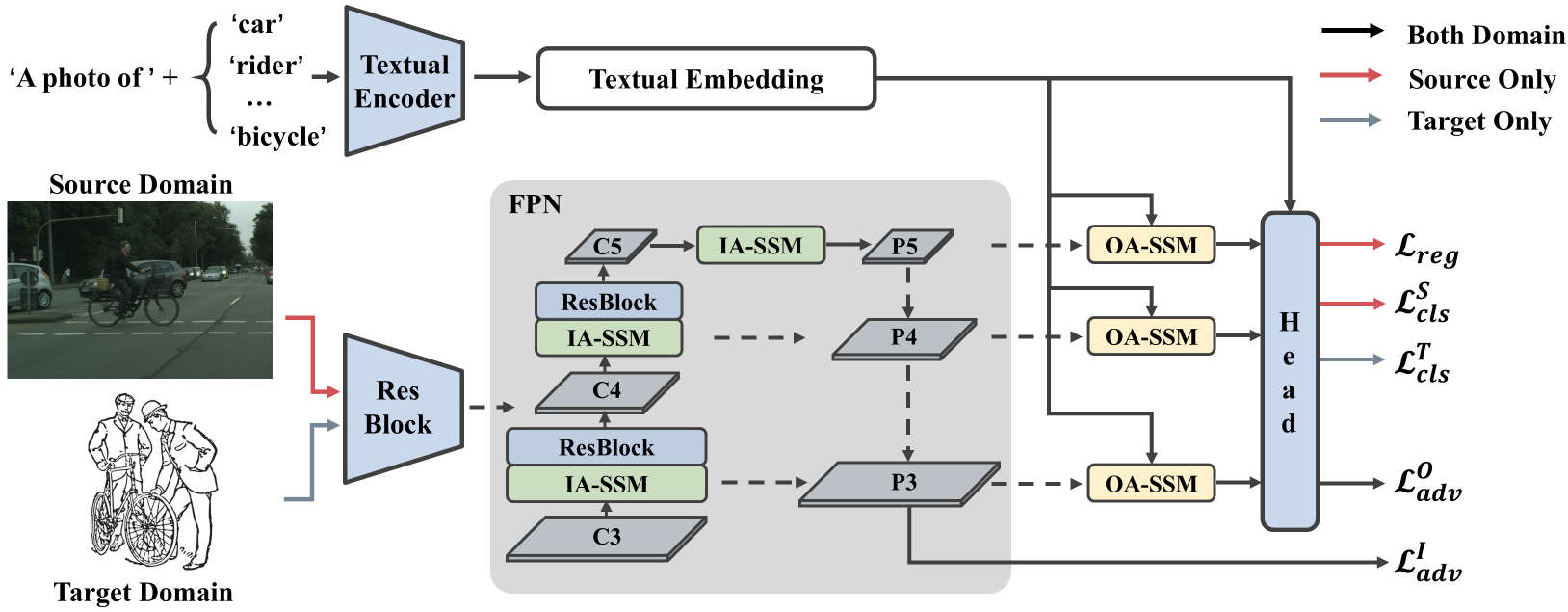} 
\vspace{-10pt}
\caption{Overview of the proposed DA-Mamba. The proposed IA-SSM and OA-SSM are integrated into the FPN of the backbone and detection head, respectively, providing fine-grained image and instance-level global-local alignment.
}
\label{fig2}
\vspace{-10pt}
\end{figure*}

In this section, we propose an innovative CNN-SSM hybrid domain adaptive object detector DA-Mamba.
DA-Mamba incorporates the Image-Aware SSM module (IA-SSM) and Object-Aware SSM module (OA-SSM) into the backbone and detection head respectively, supplementing global domain-invariant information to boost image and instance-level domain alignment.
For convenience, we implement a detector with YOLO architecture~\cite{YOLO-W} to introduce the design of DA-Mamba.
As shown in Fig.~\ref{fig2}, the input image $x$ is firstly pre-processed with a ResBlock $\mathcal{B}$ to extract low-level feature $\mathbf{C}_3$.
Then $\mathbf{C}_3$ is forwarded to FPN to extract multi-scale features, where IA-SSM $\{\mathcal{I}_i\}_{i=1}^3$ is attached to each resolution feature in the downsample stream, learning global domain information at image-level:

\begin{equation}
\begin{split}
        \mathbf{C}_4 = \mathcal{B}_1(\mathcal{I}_1(\mathbf{C}_3)); 
    \mathbf{C}_5 = \mathcal{B}_2(\mathcal{I}_2(\mathbf{C}_4));
    \mathbf{P}_5 = \mathcal{I}_3(\mathbf{C}_5),
\end{split}
\end{equation}
where $\{\mathcal{B}_j\}_{j=1}^2$ denotes ResBlock with downsample layer.
\begin{figure*}[t]
\centering
\includegraphics[width=0.95\textwidth]{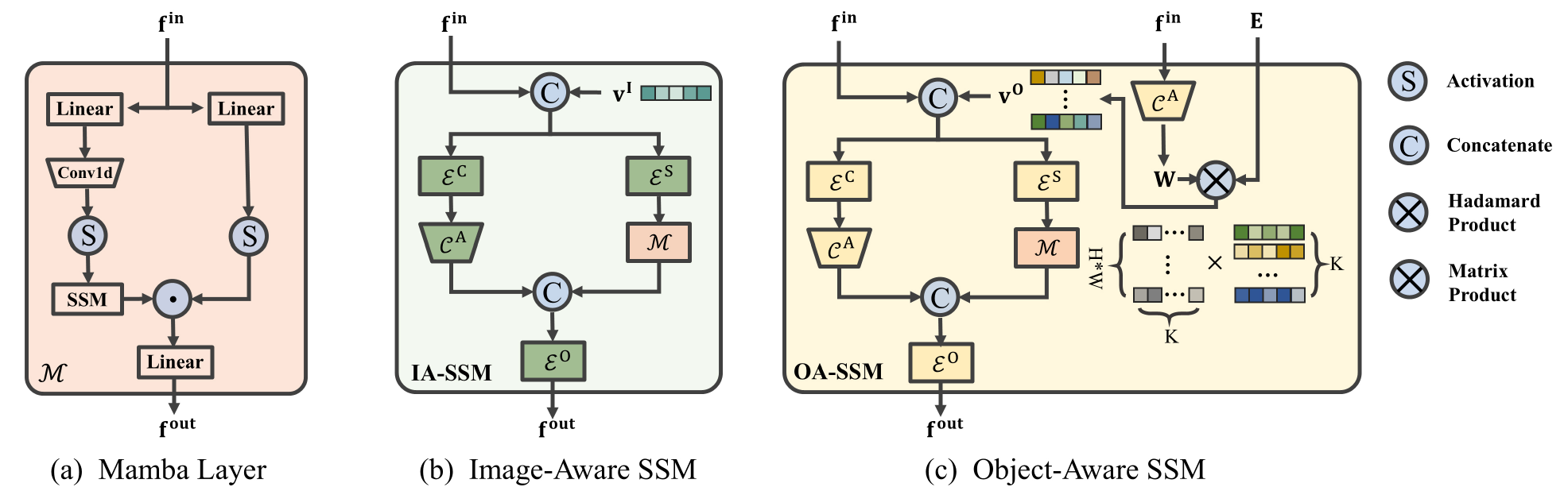} 
\vspace{-7pt}
\caption{ The architecture of (a) Mamba layer~\cite{Mamba} (b) Image-Aware SSM module and (c) Object-Aware SSM module.
}
\label{fig3}
\vspace{-13pt}
\end{figure*}

After obtaining multi-scale features $\mathbf{P}_3, \mathbf{P}_4, \mathbf{P}_5$ outputted by FPN, the OA-SSM $\{\mathcal{O}_i\}_{i=1}^3$ is also attached to each resolution feature to explore instance-level spatial and semantic dependencies in detection head $\mathcal{H}$:
\begin{equation}
    [Cls, Bbox] = \mathcal{H}(\mathcal{O}_1(\mathbf{P}_3);\mathcal{O}_2(\mathbf{P}_4);\mathcal{O}_3(\mathbf{P}_5)).
\end{equation}

\subsection{Preliminary}
State Space Models (SSMs) are mathematical frameworks widely used for modeling dynamic input sequences $\mathbf{x}(t)\in \mathbb{R}$ into output sequence $\mathbf{y}(t)\in \mathbb{R}$, with a set of hidden latent states $\mathbf{h}(t)\in \mathbb{R}^N$.
At each time step $t$, two key equations are defined to effectively model temporal dynamics:
\begin{equation}
    \label{state-update}
    \mathbf{h}'(t)=\mathbf{A}\mathbf{h}(t)+\mathbf{B}\mathbf{x}(t),
\end{equation}
\vspace{-8pt}
\begin{equation}
    \label{observation}
    \mathbf{y}(t)=\mathbf{C}^\top\mathbf{h}(t),
\end{equation}
where Eq.(\ref{state-update}) describes the updating gradient for hidden states $\mathbf{h}(t)$ with state transition matrix $\mathbf{A}\in \mathbb{R}^{N\times N}$ and state mapping matrix $\mathbf{B}\in \mathbb{R}^N$.
And the observation $\mathbf{y}(t)$ is derived from the hidden state with observation matrix $\mathbf{C}\in \mathbb{R}^N$ in Eq.(\ref{observation}).
This set of linear ordinary differential equations (ODEs) enables efficient modeling of sequential and time-dependent data while maintaining computational scalability.

In deep network architectures, data is often presented in discrete time.
To incorporate SSM into deep learning, structured state space sequence models (S4) utilize zero-hold (ZOH) method to discretize the continuous-time ODEs into a discrete-time SSM, simplifying the first-order differential terms.
Building upon the principle of S4,~\cite{Mamba} designs a unique selective discrete-time SSM, introducing attention mechanism into state transition.
Accordingly, ~\cite{Mamba} designs Mamba layer, a lightweight attention layer that operates in a low-dimensional space with linear complexity. 
As shown in Fig.~\ref{fig3}(a), Mamba combines the SSM with Gated MLP:
\begin{equation}
    \mathbf{a} = \sigma({\rm Linear}(\mathbf{f}^{in})),
\end{equation}
\vspace{-8pt}
\begin{equation}
    \mathbf{f} = {\rm SSM}(\sigma({\rm Conv1D}({\rm Linear}(\mathbf{f}^{in})))),
\end{equation}
\vspace{-8pt}
\begin{equation}
    \mathbf{f}^{out} = \mathcal{M}(\mathbf{f}^{in})={\rm Linear}(  \mathbf{a}\cdot \mathbf{f}),
\end{equation}
where $\mathbf{f}_{in}, \mathbf{f}_{out}$ is the input and output feature of Mamba layer $\mathcal{M}$, and $\sigma(\cdot)$ denotes activation function.
Considering the excellent properties of Mamba, \ie long-range perception ability with linear complexity, it provides a suitable solution for the CNN-based DAOD method to supplement the global domain information.


\subsection{Image-Aware SSM (IA-SSM)}
The Image-Aware SSM module (IA-SSM) is proposed to extract global domain information in the backbone, achieving fine-grained alignment at the image level.
Along with the CNN-based backbone to extract localized domain-invariant features, IA-SSM leverages the SSMs to explicitly capture global visual attributes in the entire image, thereby enhancing image-level domain alignment.
As shown in Fig.~\ref{fig3}(b), IA-SSM adopts a dual pipeline,\ie convolution pipeline and SSM pipeline, to extract both local and global domain information, and each IA-SSM equips a learnable image-level visual prompt $\mathbf{v}^I\in \mathbb{R}^C$ to cache image-level domain information.
Formally, the input feature $\mathbf{f}^{in}\in \mathbb{R}^{B, C, H, W}$ is firstly concatenated with image-level visual prompts broadcast from spatial locations:
\begin{equation}
    \mathbf{f} = {\rm Concat}(\mathbf{f}^{in},Broadcast(\mathbf{v}^I)).
\end{equation}

After that, feature $\mathbf{f}\in \mathbb{R}^{B,2C,H,W}$ is embedded with domain information and forwarded to the dual pipeline.
In order to effectively extract domain information and reduce redundant information in visual features, we follow the bottleneck architecture~\cite{Resnet} and reduce the dimension of input features with down-sample convolution $\mathcal{E}^S, \mathcal{E}^C$ for each pipeline:
\begin{equation}
    \mathbf{f}^S = \mathcal{E}^S(\mathbf{f}); \mathbf{f}^C = \mathcal{E}^C(\mathbf{f}),
\end{equation}
where $\mathbf{f}^S,\mathbf{f}^C \in \mathbb{R}^{B, C/r, H, W}$ and $r$ denotes reduction ratio.

In order to learn global domain-invariant features, we design the SSM pipeline, taking advantage of the Mamba layer $\mathcal{M}$ in perceiving long-range dependencies:
\begin{equation}
    \mathbf{z}^S = \mathcal{M}(\mathbf{f}^S).
\end{equation}
Meanwhile, the dual convolution pipeline is adopted to extract local domain-invariant features:
\begin{equation}
    \mathbf{z}^C=\mathcal{C}^A(\mathbf{f}^c),
\end{equation}

Finally, with the local and global domain information, we concatenate $\mathbf{z}^S$ and $\mathbf{z}^C$, mapping them back to the original feature space with up-sample convolution $\mathcal{E}^O$:
\begin{equation}
    \mathbf{f}^{out}=\mathcal{E}^O({\rm Concat(\mathbf{z}^S,\mathbf{z}^C)}),
\end{equation}
where $\mathbf{f}^{out}\in \mathbb{R}^{B,C,H,W}$ is the output of IA-SSM.

By integrating IA-SSM, we enable the detection backbone with the ability to learn both local and global domain-invariant features, facilitating domain alignment at the image level.

\subsection{Object-Aware SSM (OA-SSM)}
The Object-Aware SSM module (OA-SSM) is introduced into the detection head, modeling the spatial and semantic dependencies between all instances to align global domain information at the instance-level.
As shown in Fig.~\ref{fig3}(c), in order to extract local and global domain information in the detection head effectively, OA-SSM shares a similar network architecture as IA-SSM.
However, unlike the image-level visual prompt $\mathbf{v}^I$ in IA-SSM, an ideal visual prompt for OA-SSM needs to satisfy the following design points: 1) to learn spatial dependencies, it needs to be pixel-wise; 2) to learn semantic dependencies, it should be conditioned on the input features and interact with the semantics of various categories.

To achieve this, we propose to map the visual features into a pixel-wise category similarity matrix with a learnable meta-net, and multiply it with prototypes of all categories to obtain the instance-level visual prompt $\mathbf{v}^O\in \mathbb{R}^{B,C,H,W}$.
Formally, given the input feature $\mathbf{f}^{in}\in \mathbb{R}^{B, C, H, W}$, we project it into a category similarity matrix $\mathbf{W}\in \mathbb{R}^{B,H,W,K}$, where $K$ is the total number of categories.
This projection is implemented using a convolutional layer $\mathbf{C}^A$ with reshape operation.
Subsequently, the instance-level visual prompt $\mathbf{v}^O$ is defined as the matrix product of $\mathbf{W}$ and category prototype $\mathbf{E}\in \mathbb{R}^{K, C}$:
\begin{equation}
    \mathbf{v}^O = \mathbf{W}\otimes\mathbf{E}=\mathcal{C}^A(\mathbf{f}^{in})\mathbf{E},
\end{equation}
where category prototype $\mathbf{E}$ acts as global semantic anchors, describing the relative distances of visual features to each category.
Leveraging the alignment capability of vision-language models (VLMs), we utilize the textual encoder of CLIP to generate category prototypes. 
Specifically, we input textual prompts such as “A photo of [Class]” to represent each category, ensuring a consistent mapping between visual features and semantic categories.

Since a single instance presents a defined semantics, all pixels in the same instance region share the same visual prompt. 
Therefore, $\mathbf{v}^O$ divides the feature map into instance-level semantic regions, \ie, each region shares consistent category semantics, and regions have different category semantics. 
After generating $\mathbf{v}^O$, OA-SSM will concatenate it with the input features $\mathbf{f}^{in}$ and forward to the dual pipeline, modeling instance-level spatial and semantic dependencies in the detection head.


\subsection{Optimization Objective}
\label{loss}
We propose DA-Mamba by introducing IA-SSM and OA-SSM into detection backbone and head to learn domain-invariant and task-relevant knowledge.
Therefore, the optimization object contains four sub-items: image-level adversarial loss, instance-level adversarial loss, source domain supervised loss and targe domain semi-supervised loss.

To ensure IA-SSM learning image-level domain-invariant knowledge, we constrain the output of IA-SSM $\mathbf{f}_{out}$ to be domain-indistinguishable.
Specifically, each IA-SSM is followed by a domain discriminator $\{\mathcal{D}^I_i\}_{i=1}^3$ with Gradient Reverse Layer(GRL)~\cite{DA-Faster} to calculate image-level adversarial loss $\mathcal{L}^I_{adv}$:
\begin{equation}
        \mathcal{L}^I_{adv}=-\sum_{i=1}^3[\mathbb{E}_{\mathbf{x}_s}||\mathcal{D}^I_i(\mathbf{f}^{out}_i)||_2^2+\mathbb{E}_{\mathbf{x}_t}||\mathcal{D}^I_i(\mathbf{f}^{out}_i)-\mathbf{1}||_2^2].
\end{equation}

Meanwhile, each OA-SSM is also followed by discriminators $\{\mathcal{D}^O_i\}_{i=1}^3$ with mask to calculate object-level adversarial loss $\mathcal{L}_{o\_adv}$:
\begin{equation}
\begin{split}
        \mathcal{L}^O_{adv}=-\sum_{i=1}^3[&\mathbb{E}_{\mathbf{x}_s}||\mathbf{M}\cdot\mathcal{D}^O_i(\mathbf{f}^{out}_i)||_2^2 \\+&\mathbb{E}_{\mathbf{x}_t}||\mathbf{M}\cdot(\mathcal{D}^O_i(\mathbf{f}^{out}_i)-\mathbf{1})||_2^2].
\end{split}
\end{equation}
where $\mathbf{M}$ is an instance-level mask.
$\mathbf{M}$ is generated by filtering regions where the classification probability exceeds $0.5$, constraining OA-SSM to adaptively focus on the relationship between foreground instance regions rather than the relationship between each pixel. 

For the source domain, we minimize the cross-entropy $\mathcal{L}^S_{cls}$ and regression loss $\mathcal{L}_{reg}$.
For the target domain, we select predictions with high confidence as pseudo labels and minimize the cross-entropy $\mathcal{L}^T_{cls}$.
Overall, the optimization objective is:
\begin{equation}
    \mathcal{L}=\mathcal{L}^S_{cls} + \mathcal{L}^T_{lcs} + \lambda^I\mathcal{L}^I_{adv}+\lambda^O\mathcal{L}^O_{adv} + \mathcal{L}_{reg}
\end{equation}

\section{Experiment}

\subsection{Dataset}
\label{Dataset}
\noindent\textbf{Cross-Weather} 
Cityscapes~\cite{Cityscapes} contains diverse street scenes in daylight, consisting of 2,975 training and 500 validation images annotated with eight classes. 
Foggy Cityscapes~\cite{FoggyCityscapes} simulates three distinct densities of fog on Cityscapes, containing 8,925 training images and 1,500 validation images. 
A standard configuration for cross-weather adaptation is to take the training set of Cityscapes as source domain and the training set of foggy Cityscapes as target domain, evaluating cross-weather adaptation performance on the 1500-sized validation set in all eight categories.

\noindent\textbf{Cross-FoV}
BDD100K~\cite{bdd} is crucial for self-driving, including 36,278 training images and 5,258 validation images with diverse scenarios and Field of View (FOV) in the daytime subset. 
To fairly compare with other methods, we migrate Cityscapes to BDD100k on $7$ shared categories.

\begin{table*}[ht]
\centering
\caption{Comparison ($\%$) on Cross-Weather adaptation task Cityscapes$\rightarrow$Foggy Cityscapes (C$\rightarrow$F)}
\label{tab1}
\setlength\tabcolsep{10.5pt}
\vspace{-8pt}
\resizebox{1.0\textwidth}{!}{ 
\begin{tabular}{ccccccccccc}
\toprule
   & &\multicolumn{9}{c}{C$\rightarrow$F}\\
    \cmidrule(r){3-11} 
Methods&Venue&Person&Rider&Car&Truck&Bus&Train&Motor&Bicycle&mAP\\
\midrule
DA-Faster~\cite{DA-Faster}& CVPR 2018 &29.2 &40.4 &43.4 &19.7 &38.3 &28.5 &23.7 &32.7 &32.0 \\
LRA~\cite{LRA}& TNNLS 2024  &45.6& 47.1& 59.7& 31.2 &52.4 &44.6 &28.1 &39.5 &43.5\\
SIGMA++~\cite{sigma++}& TPAMI 2023 & 46.4 & 45.1 & 61.0 & 32.1 & 52.2 & 44.6 & 34.8 & 39.9 & 44.5\\
FA-TDCA~\cite{FA-TDCA}& TNNLS 2025 &43.5 &51.7 &58.8 &34.1 &52.6 &40.4 &36.0 &42.3 &44.9 \\
MTM~\cite{MTM}&  AAAI 2024  &51.0 &53.4 &67.2 &37.2 &54.4 &41.6 &38.4 &47.7 &48.9\\
AT~\cite{AT}& CVPR 2022 &56.3 &51.9 &64.2 &38.5 &45.5 &\textbf{55.1} &\textbf{54.3} &35.0 &50.9\\
SOCCER~\cite{SOCCER}& ACMMM 2024  &51.7 &57.7 &68.6 &38.2 &51.6 &47.5 &41.6 &51.7 &51.1\\
MRT~\cite{MRT}& ICCV 2023  &52.8 &51.7 &68.7 &35.9 &58.1 &\underline{54.5} &41.0 &47.1 &51.2\\
DSD-DA~\cite{DSD-DA}&  ICML 2024      &49.0 &59.6 &65.3 &35.7 &61.0 &46.5 &43.9 &57.3 &52.3\\
CAT~\cite{CAT}&  CVPR 2024   &44.6 &57.1 &63.7 &\underline{40.8} &66.0 &49.7 &44.9 &53.0 &52.5 \\
NSA-UDA~\cite{NSA}& ICCV 2023  &50.3 &60.1 &67.7 &37.4 &57.4 &46.9 &47.3 &54.3 &52.7\\
REACT~\cite{REACT}& TIP 2024 &51.4 &57.9 &67.4 &37.7 &58.4 &52.8 &44.6 &54.6 &53.1 \\
DATR~\cite{DATR} & TIP 2024 &\textbf{61.6} &60.4 &74.3 &35.7 &60.3 &35.4 &43.6 &55.9 &53.4 \\
DT~\cite{DT} & CVPR 2025 & 48.5 & 60.0 & 65.4 & 47.2 & \textbf{66.5} & 52.9 & 46.2 & 56.7 & 55.4 \\
DA-Pro~\cite{DA-Pro}& NeurIPS 2023 &{55.4} &\textbf{62.9} &\underline{70.9} &40.3 &{63.4} &54.0 &42.3 &\textbf{58.0} &\underline{55.9}\\
\midrule
Baseline(UDA)& CVPR 2024 &57.5 & 45.9 & 68.5 & 38.4& 61.2& 50.9& 40.8& 55.5 & 52.3 \\
DA-Mamba (Ours)& - &\underline{59.3} &\underline{61.5} &\textbf{71.0} &\textbf{43.5} &\underline{66.2} &\textbf{55.1} &\underline{50.1} &\underline{57.8} &\textbf{58.1}\\
\midrule
Oracle& - &61.8 & 69.0 & 72.4 & 46.8& 69.5& 57.6& 52.7& 58.9&61.1\\
\bottomrule
\end{tabular}  
}
\vspace{-15pt}
\end{table*}

\noindent\textbf{Cross-Style} Pascal VOC~\cite{pascal} dataset consists of 16551 images from the real world and has 20 categories.
Comic~\cite{comic} contains 1,000 training images and 1,000 test images, rendered in artistic styles, and share 6 categories with Pascal VOC.
Clipart~\cite{clipart} contains 1000 comical images and has the same 20 categories as Pascal VOC.
Following the mainstream splitting, we use Pascal VOC as the source domain and all Clipart images as the target domain. 
For Pascal VOC$\rightarrow$Comic, we use the training split for training and the testing split for evaluation.

\subsection{Implementation Details}
\label{Implementation}
We introduce the one-stage detector YOLO-World~\cite{YOLO-W} into the unsupervised domain adaptation (UDA) setting as the baseline detector.
In each iteration, one batch of source images with ground truth and one batch of unlabeled target domain images are forwarded to calculate the classification, adversarial and regression loss.
The hyperparameter $r, \lambda^{I}, \lambda^{O}$ is set to $2.0, 1.0, 0.5$.
The batch size for each domain is set to $2$, using the SGD optimizer with linear learning rate decay.
Mean Average Precision (mAP) with a threshold of 0.5 is taken as the evaluation metric. 
All experiments are deployed on 1 Tesla V100 GPUs.

\begin{table*}[t]
\centering
\caption{Comparison ($\%$) on Cross-FOV adaptation task Cityscapes$\rightarrow$BDD100K (C$\rightarrow$B) and Cross-Style adaptation task Pascal VOC$\rightarrow$Comic (P$\rightarrow$Cmc)}
\label{tab3}
\vspace{-10pt}
\setlength\tabcolsep{6pt}
\resizebox{1.0\textwidth}{!}{      
\begin{tabular}{cccccccccccccccc}
\toprule
   &\multicolumn{8}{c}{C$\rightarrow$B}&\multicolumn{7}{c}{P$\rightarrow$Cmc}\\
    \cmidrule(r){2-9}     \cmidrule(r){10-16} 

Methods& Person & Rider&Car&{Truck}&Bus&Motor&Bicycle&mAP&{Bicycle} & {Bird}&{Car}&{Cat}&{Dog}&{Person}&mAP\\ 
\midrule
DA-Faster~\cite{DA-Faster} &28.9 &27.4 &44.2 &19.1 &18.0 &14.2 &22.4 &24.9&-&-&-&-&-&-&- \\
SWDA~\cite{Strong-weak}& 30.2& 29.5& 45.7& 15.2& 18.4& 17.1& 21.2& 25.3& 36.0& 18.3& 29.3& 9.3& 22.9& 48.4& 27.4\\
DBGL~\cite{DBGL}&-&-&-&-&-&-&-&-&35.6&20.3&33.9&16.4&26.6&45.3&29.7\\
FGRR~\cite{FGRR}&-&-&-&-&-&-&-&-&42.2&21.1&30.2&21.9&30.0&51.5&32.7\\
SFA~\cite{SFA}& 40.2 &27.6 &57.5 &19.1 &23.4 &15.4 &19.2 &28.9&-&-&-&-&-&-&-\\
BiADT~\cite{BiADT}&42.0 &34.5 &59.9 &17.2 &19.2 &17.8 &24.4 &32.7&-&-&-&-&-&-&-\\
TDD~\cite{TDD}& 39.6& 38.9& 53.9& 24.1& 25.5 &24.5 &28.8& 33.6&-&-&-&-&-&-&-\\
MRT~\cite{MRT}& 48.4 &30.9 &63.7 &24.7 &25.5 &20.2 &22.6 &33.7&-&-&-&-&-&-&-\\
SIGMA++~\cite{sigma++} & 47.5 &30.4 &65.6 &21.1 &26.3 &17.8 &27.1 &33.7 & 45.7 &22.4 &36.9 &25.5 &\underline{27.9} &62.5 &37.1\\
D-adapt~\cite{d-adapt}   &-&-&-&-&-&-&-&-&52.4 &\textbf{25.4} &42.3 &\textbf{43.7} &25.7 &53.5 &\underline{40.5} \\
DATR~\cite{DATR} &\underline{58.5} &\textbf{42.8} &\textbf{73.4} &26.9 &39.9 &24.2 &\underline{37.3} &\underline{43.3}&-&-&-&-&-&-&-\\
\midrule
Baseline(UDA)& 54.5& 25.0& 66.0& \underline{37.5}& \underline{40.7}&\underline{33.0}&36.9&41.9&\underline{60.2}&17.8&\underline{52.3}&15.3&17.8&\underline{63.9}&37.9  \\
DA-Mamba(Ours)& \textbf{63.5}& \underline{41.6}& \underline{71.4}& \textbf{43.1}& \textbf{42.3}& \textbf{41.0}& \textbf{38.1}& \textbf{48.7}& \textbf{61.7}& \underline{23.1}& \textbf{55.9}& \underline{27.2}& \textbf{28.1}& \textbf{66.6}& \textbf{43.8} \\
\midrule
Oracle& 67.5& 45.4& 77.3& 51.0& 51.8& 44.6& 43.4& 54.4
& 66.7& 27.1& 64.0& 27.9& 32.8& 70.8& 48.2  \\

\bottomrule
\end{tabular}  
}
\vspace{-10pt}
\end{table*}

\begin{table*}[t]
\centering
\caption{Comparison ($\%$) on Cross-Style adaptation task Pascal VOC$\rightarrow$Clipart (P$\rightarrow$Clp)}
\label{tab2}
\vspace{-10pt}
\setlength\tabcolsep{4pt}
\resizebox{1.0\textwidth}{!}{      
\begin{tabular}{cccccccccccccccccccccc}
\toprule
Methods& \rotatebox{90}{Aero} & \rotatebox{90}{Bike}&\rotatebox{90}{Bird}&\rotatebox{90}{Boat}&\rotatebox{90}{Bottle}&\rotatebox{90}{Bus}&\rotatebox{90}{Car}&\rotatebox{90}{Cat}&\rotatebox{90}{Chair}&\rotatebox{90}{Cow}&\rotatebox{90}{Table}&\rotatebox{90}{Dog}&\rotatebox{90}{Horse}&\rotatebox{90}{Motor}&\rotatebox{90}{Person}&\rotatebox{90}{Plant}&\rotatebox{90}{Sheep}&\rotatebox{90}{Sofa}&\rotatebox{90}{Train}&\rotatebox{90}{Tv}&mAP\\ 
\midrule
UaDAN~\cite{UaDAN} & 35.0 & \underline{73.7} & \underline{41.0} & 24.4 & 21.3 & \underline{69.8} & 53.5 & 2.3 & 34.2 & 61.2 & 31.0 & \textbf{29.5} & 47.9 & 63.6 & 62.2 & \underline{61.3} & 13.9 & 7.6 & 48.6 & 23.9 & 40.2 \\
TFD~\cite{TFD} & 27.9 & 64.8 & 28.4 & 29.5 & 25.7 & 64.2 & 47.7 & 13.5 & 47.5 & 50.9 & 50.8 & 21.3 & 33.9 & 60.2 & 65.6 & 42.5 & 15.1 & 40.5 & 45.5 & 48.6 & 41.2 \\
UMT~\cite{UMT} & 39.6 & 59.1 & 32.4 & 35.0 & 45.1 & 61.9 & 48.4 & 7.5 & 46.0 & \textbf{67.6} & 21.4 & \textbf{29.5} & 48.2 & 75.9 & 70.5 & 56.7 & 25.9 & 28.9 & 39.4 & 43.6 & 44.1 \\
ATMT~\cite{ATMT}  & 37.5 & 63.4 & 37.9 & 29.8 & 45.1 & 62.7 & 41.2 & 19.5 & 43.7 & 57.4 & 22.9 & 25.3 & 39.6 & 87.1 & 70.9 & 50.6 & 29.1 & 32.2 & \underline{58.4} & 50.5 & 45.2 \\
CIGAR~\cite{CIGAR} &35.2 &55.0 &39.2 &30.7 &60.1 &58.1 &46.9 &\underline{31.8} &47.0 &61.0 &21.8 &26.7 &44.6 &52.4 &68.5 &54.4 &31.3 &38.8 &56.5 &\textbf{63.5} &46.2\\
TIA~\cite{TIA} & \underline{42.2} & 66.0 & 36.9 & 37.3 & 43.7 & \textbf{71.8} & 49.7 & 18.2 & 44.9 & 58.9 & 18.2 & \underline{29.1} & 40.7 & \underline{87.8} & 67.4 & 49.7 & 27.4 & 27.8 & 57.1 & 50.6 & 46.3 \\
SIGMA++~\cite{sigma++} & 36.3 & 54.6 & 40.1 & 31.6 & 58.0 & 60.4 & 46.2 & \textbf{33.6} & 44.4 & \underline{66.2} & 25.7 & 25.3 & 44.4 & 58.8 & 64.8 & 55.4 & \underline{36.2} & 38.6 & 54.1 & \underline{59.3} & 46.7 \\
CMT~\cite{CMT} & 39.8 & 56.3 & 38.7 & 39.7 & 60.4 & 35.0 & \underline{56.0} & 7.1 & 60.1 & 60.4 & 35.8 & 28.1 & \textbf{67.8} & 84.5 & \textbf{80.1} & 55.5 & 20.3 & 32.8 & 42.3 & 38.2 & 47.0 \\
DA-Ada~\cite{DA-Ada} &42.3 &\textbf{75.1} &48.9 &\underline{45.9} &49.0 &\textbf{71.8} &55.6 &15.4 &50.7 &56.6 &19.9 &20.6 &61.3 &80.7 &\underline{73.0} &29.2 &\textbf{37.5} &21.5 &52.5 &52.9 &48.0\\

CAT~\cite{CAT} &40.5 &64.1 &38.8 &41.0 &60.7 &55.5 &55.6 &14.3 &54.7 &59.6 &46.2 &20.3 &\underline{58.7} &\textbf{92.9} &62.6 &57.5 &22.4& 40.9 &49.5 &46.0 &\underline{49.1}\\

\midrule
Baseline(UDA) & 38.1& 61.1& \underline{41.0}& 38.8& \underline{63.7}& 51.1& 53.2& 17.9& \underline{66.6}& 24.9& \underline{56.0}& 15.8&44.8& 64.1& 60.7& 53.9&22.7&\underline{53.3}&53.0&43.2&46.2\\
DA-Mamba(Ours) & \textbf{46.7} & 68.8 & \textbf{42.9} & \textbf{51.3} & \textbf{69.1} & \textbf{53.1} & \textbf{59.0} & 24.1 & \textbf{70.0} & 33.8 & \textbf{59.5} & 20.5 & 48.5 & 68.2 & 70.9 & \textbf{67.2} & 25.6 & \textbf{60.7} & \textbf{58.5} & 50.8 & \textbf{52.5} \\
\midrule
Oracle & 46.2& 71.3& 45.3& 52.5& 72.0& 53.1& 60.9& 25.5& 72.0& 34.8& 60.1& 23.8& 50.6& 63.6& 72.8& 67.7& 28.3& 63.7&59.8&51.5&53.8\\

\bottomrule
\end{tabular}  
}
\vspace{-15pt}
\end{table*}

\subsection{Comparison to SOTA methods}
We present representative DAOD approaches for comparison, including both CNN and Transformer-based detectors.

\noindent\textbf{Cross-Weather Adaptation Scenario}
As shown in Table~\ref{tab1}, the proposed DA-Mamba outperforms all CNN or Transformer-based methods and advances SOTA by $2.2\%(55.9\% \rightarrow 58.1\%)$ mAP.
Specifically, our method improves performance on all categories over baseline, ranging from $1.8\%$ to $15.6\%$, and achieves a significant total improvement of $5.8\%$, close to the oracle $61.1\%$.

\noindent\textbf{Cross-FoV Adaptation Scenario}
Table~\ref{tab3} (C$\rightarrow$B) shows that the proposed method achieves the best results of $48.7\%$ mAP and outperforms the SOTA DATR~\cite{DATR} $43.3\%$ with a large margin of $5.4\%$, showing the effectiveness of DA-Mamba in extracting spatial and semantic dependencies.

\noindent\textbf{Cross-Style Adaptation Scenario}
Additionally, we assess DA-Mamba on the more challenging Cross-Style adaptation, where the semantic hierarchy has broader discrepancies. 
In Table~\ref{tab3} (P$\rightarrow$Cmc), DA-Mamba peaks with $43.8\%$ mAP, surpassing the SOTA D-adapt~\cite{d-adapt} by a large margin of $3.3\%$.
In Table~\ref{tab2}, DA-Mamba achieves a significant improvement of $3.4\%$ over the SOTA CAT~\cite{CAT}, only slightly lower than the oracle by $1.3\%$.

\subsection{Ablation Studies}

\begin{table}[t]
\centering
\large
\caption{\small Ablation ($\%$) on DA-Mamba. }
\label{tab-component}
\setlength\tabcolsep{5pt}
\vspace{-9pt}
\resizebox{1.0\columnwidth}{!}{      
\begin{tabular}{ccccc}
\toprule
Method & C$\rightarrow$F & C$\rightarrow$B & P$\rightarrow$Clp & P$\rightarrow$Cmc\\ 
  \midrule
w/o          &  52.3 & 41.9 & 46.8 & 37.9    \\
w/ IA-SSM &  56.8(+4.5) & 45.9(+4.0) & 50.6(+3.8) & 40.5(+2.6)    \\
w/ OA-SSM &  55.4(+3.1) & 45.0(+3.1) & 50.5(+3.7) & 40.6(+2.7)    \\
w/ both   &  58.1(+5.8) & 48.7(+6.8) & 52.5(+5.7) & 43.8(+5.9)    \\
\bottomrule
\end{tabular}  }  
\vspace{-10pt}
\end{table}

\begin{table}[t]
\centering
\caption{\small Comparison ($\%$) with standard Mamba. 
}
\label{tab-mamba}
\setlength\tabcolsep{5pt}
\vspace{-9pt}
\resizebox{1.0\columnwidth}{!}{      
\begin{tabular}{cccccc}
\toprule
Component&Method & C$\rightarrow$F  & C$\rightarrow$B &P$\rightarrow$Clp &P$\rightarrow$Cmc\\ 
\midrule
\multirow{2}{*}{Backbone}&Mamba~\cite{U-Mamba} &54.3 & 43.4&48.3  & 38.7 \\ 
&IA-SSM         & 56.8 & 45.9 & 50.6 & 40.5  \\
\midrule
\multirow{2}{*}{Head}&Mamba~\cite{U-Mamba} &53.4  &43.3  & 47.6&38.4  \\ 
&OA-SSM         & 55.4 & 45.0 & 50.5 & 40.6  \\
\midrule
\multirow{2}{*}{Backbone+Head}&Mamba~\cite{U-Mamba} &55.0  &44.8  & 48.7&38.9  \\ 
&DA-Mamba         & 58.1 & 48.7 & 52.5 & 43.8  \\
\bottomrule
\end{tabular}   }
\vspace{-15pt}
\end{table}

\noindent\textbf{IA-SSM and OA-SSM} \textit{IA-SSM and OA-SSM improve the domain alignment capability from orthogonal perspectives.}
As shown in Table~\ref{tab-component}, introducing IA-SSM achieves $2.6\sim 4.5\%$ mAP improvements over the baseline, and attaching OA-SSM also gains $2.7\sim 3.7\%$.
With both modules, our DA-Mamba achieves total improvements of $5.7\sim 6.8\%$, showing that it achieves fine-grained domain alignment in the backbone and head, respectively.

\noindent\textbf{Standard Mamba \textit{vs.} DA-Mamba.} \textit{DA-Mamba achieves better learning of domain-invariant features by balancing local and global alignment.}
We replace the IA-SSM in the backbone and OA-SSM in the detection head with standard Mamba layers~\cite{U-Mamba}.
As shown in Table~\ref{tab-mamba}, IA-SSM surpasses the vanilla Mamba by $1.8\sim2.5\%$, indicating that directly applying Mamba to DAOD is suboptimal.
Similarly, OA-SSM outperforms Mamba by $1.7\sim2.9\%$ in the detection head, showing superiority in modeling instance-level spatial and semantic relationships across domains.
Applied to both image and instance-level, vanilla Mamba layer only brings limited improvements ($1.0\sim2.9\%$), while DA-Mamba improves $5.7\sim6.8\%$ significantly.
The reason for the limited adaptation performance of standard Mamba lies in its domain-agnostic state update strategy, which fails to align long-range dependencies when the feature distributions between domains diverge.
In contrast, IA-SSM and OA-SSM leverage convolutional local priors as guidance, ensuring that global dependencies are aligned with locally stable domain-invariant features.
This conclusion is further verified under larger domain discrepancies: when adapting under cross-style scenarios P$\rightarrow$Clp and P$\rightarrow$Cmc, standard Mamba achieves only marginal gains ($1.9\%$ and $1.0\%$), whereas DA-Mamba maintains substantial improvements ($5.7\%$ and $5.9\%$).
Overall, DA-Mamba inherits the global modeling strength of Mamba while grounding it in convolutional local priors, achieving robust domain-invariant representation learning under domain shift.
\vspace{-6pt}

\begin{table*}[h]
\centering
\caption{Comparison ($\%$) of computational overhead.
}
\label{tab-computational}
\setlength\tabcolsep{3pt}
\vspace{-8pt}
\resizebox{1.0\textwidth}{!}{      
\begin{tabular}{ccccccccc}
\toprule
 Method  & Arch. & Inference Mem. Usg. & FPS & Training Mem. Usg. & Training s/iter & FLOPs & mAP & Abs. Gains\\
\midrule
DA-Pro~\cite{DA-Pro}& Visual-Language Model       & 2549M & 2.5  &  \textbf{4034M} & 1.472S & 243G & 55.9& 3.3\\
DATR~\cite{DATR}    & Transformer-based Attention & 3206M & 5.8  & 21606M & 0.906s & 279G & 53.4& \underline{5.3}\\
\midrule
Baseline            & CNN-based Backbone          & \textbf{1190M}  & \textbf{15.2} & \underline{14581M} & \textbf{0.703s} & \textbf{140G} & 52.3& - \\ 
Baseline+GC Block   & CNN-based Attention         & 1341M  & \underline{14.1} & 15238M  & \underline{0.776s} & \underline{146G} & 54.1& 1.8\\ 
Baseline+ViT Block  & Transformer-based Attention & 2090M  & 10.6 & 19322M & 0.862s & 208G & \underline{57.5}& 5.2\\ 
Baseline+Mamba Block& SSM-based Attention         & 1792M  & 13.0 & 18027M & 0.799s & 182G &  55.0& 2.7\\ 
\midrule
DA-Mamba(Ours)      & CNN-SSM Hybrid Attention & \underline{1307M} & \underline{14.1} & 15500M & 0.782s & 148G & \textbf{58.1}& \textbf{5.8}\\
\bottomrule
\end{tabular}   }
\vspace{-15pt}
\end{table*}

\begin{table}[t]
    \centering
\caption{\small Ablation ($\%$) on IA-SSM. }
\label{tab-IA-SSM}
\setlength\tabcolsep{6pt}
\vspace{-8pt}
\resizebox{0.98\columnwidth}{!}{      
\begin{tabular}{ccccccc}
\toprule
Conv & SSM & Prompts & C$\rightarrow$F  & C$\rightarrow$B &P$\rightarrow$Clp &P$\rightarrow$Cmc\\ 
\midrule
           &             &             & 52.3 & 41.9 & 46.8 & 37.9  \\
\checkmark &             &             & 53.5 & 43.8 & 47.6 & 38.3  \\
           & \checkmark  &             & 54.3 & 43,4 & 48.3 & 38.7  \\
\checkmark & \checkmark  &             & 55.6 & 45.0 & 49.2 & 40.0  \\
\checkmark &             &  \checkmark & 53.9 & 44.0 & 48.3 & 38.4  \\
           & \checkmark  &  \checkmark & 54.9 & 43.9 & 49.0 & 40.0  \\
\checkmark & \checkmark  & \checkmark  & 56.8 & 45.9 & 50.6 & 40.5  \\

\bottomrule
\end{tabular}  }  
\vspace{-10pt}
\end{table}

\begin{table}[t]
    \centering
\caption{\small Ablation ($\%$) on OA-SSM. 
}
\label{tab-OA-SSM}
\setlength\tabcolsep{6pt}
\vspace{-8pt}
\resizebox{0.98\columnwidth}{!}{      
\begin{tabular}{ccccccc}
\toprule
Conv & SSM & Prompts & C$\rightarrow$F  & C$\rightarrow$B &P$\rightarrow$Clp &P$\rightarrow$Cmc\\ 
\midrule
           &             &             & 52.3 & 41.9 & 46.8 & 37.9  \\
\checkmark &             &             & 53.2 & 43.2 & 47.5 & 38.1  \\
           & \checkmark  &             & 53.4 & 43.3 & 47.6 & 38.4  \\
\checkmark & \checkmark  &             & 54.8 & 44.6 & 49.4 & 39.8  \\
\checkmark &             &  \checkmark & 53.8 & 43.3 & 47.8 & 38.5  \\
           & \checkmark  &  \checkmark & 54.5 & 44.0 & 48.7 & 39.2  \\
\checkmark & \checkmark  & \checkmark  & 55.4 & 45.0 & 50.5 & 40.6  \\
\bottomrule
\end{tabular}   }
\vspace{-15pt}
\end{table}


\noindent\textbf{Computational Overhead.} \textit{DA-Mamba achieves superior performance with minimal additional cost.}
As shown in Table~\ref{tab-computational}, compared with SOTA method DA-Pro\cite{DA-Pro}, DA-Mamba gains mAP improvements of $2.2\%$ with $51.2\%$ inference memory usage, $5.64\times$FPS and $57.6\%$ FLOPs.
Compared with transformer-based DATR~\cite{DATR}, DA-Mamba attains $4.7\%$ higher mAP, $2.62\times$FPS with merely $37.1\%$ inference memory, $67.5\%$ training memory and $50.2\%$ FLOPs, highlighting efficiency in both speed and accuracy through linear attention.
Within the same CNN-based framework, the plain baseline yields the highest efficiency (15.2 FPS, 140G FLOPs) but suffers from the lowest mAP ($52.3\%$) due to the lack of global alignment.
Introducing the GC Block~\cite{GCNet} to model global dependencies with convolutional attention variants yields a modest $1.8\%$ mAP gain with slight memory and FLOPs increases, yet remains constrained by its globally shared attention weights.
Replacing it with Transformer block substantially boosts global perception and raises mAP to $57.5\%$, though at the cost of $1.49\times$ FLOPs, $69.7\%$ FPS, $1.76\times$ inference and $1.33\times$ training memory usage.
In comparison, the Mamba layer achieves a balanced trade-off between performance and computational overhead.
Finally, DA-Mamba integrates IA-SSM and OA-SSM into the CNN backbone and head, maintaining near-baseline efficiency (14.1 FPS, 148G FLOPs) while achieving the best $58.1\%$ mAP and $5.8\%$ gains.
This demonstrates that DA-Mamba effectively combines lightweight efficiency of CNNs with linear global modeling capability of SSMs, achieving global–local domain alignment without heavy computation of Transformers or limited expressiveness of convolutional attention.

\noindent\textbf{Pipeline and Prompts.}  \textit{Both the convolution pipeline and the SSM pipeline are indispensable for effectively combining local and global features, while visual prompts further enhance domain information.}
As shown in Table~\ref{tab-IA-SSM} and Table~\ref{tab-OA-SSM}, introducing the convolution pipeline yields $0.4\sim 1.9\%$ improvements for IA-SSM and $0.2\sim1.3\%$ for OA-SSM, as it provides vital local domain-invariant features. 
And the SSM pipeline brings $0.8\sim2.0\%$ and $0.5\sim1.4\%$ gains, benefiting from its ability to capture long-range dependencies across domains.
With both pipelines, IA-SSM and OA-SSM achieve the highest improvements of $2.1\sim 3.3\%$ and $1.9\sim 2.6\%$, confirming that local and global alignment are complementary and jointly necessary for robust domain-invariant representation learning.
Meanwhile, image-level and instance-level visual prompts further boost the performance with $0.5\sim1.4\%$ and $0.4\sim1.1\%$, since prompts condition the dual pipeline on domain and category-specific context, enabling better domain alignment at both image and instance-level.
Overall, results demonstrate that DA-MAmba benefits from a synergistic integration of convolutional locality, SSM-based global perception and prompt-guided semantic adaptation, outperforming existing methods that rely on any single component.

\noindent\textbf{Insertion Sites.} \textit{SSM can improve the alignment of features at all levels and is more effective for high-level features.}
We study the insertion site of IA-SSM and OA-SSM, as shown in Table~\ref{tab-insertion}.
For IA-SSM in the backbone, attaching it to the high-level feature $\mathbf{C}_5$ ($+2.0\%$) is more effective than low-level feature $\mathbf{C}_3$ ($+1.5\%$).
We attribute this to the fact that high-level features contain more semantic information rather than texture, which is more suitable for IA-SSM to analyze global domain information.
For the detection head, OA-SSM presents similar performance ($+1.6\sim1.8\%$) on $\mathbf{P}_3, \mathbf{P}_4, \mathbf{P}_5$, because the feature fusion of FPN makes them differ mainly in scale rather than semantics.

\begin{table}[t]
\centering
\caption{Ablation studies ($\%$) for insertion sites of DA-Mamba.}
\label{tab-insertion}
\setlength\tabcolsep{11pt}
\vspace{-8pt}
\resizebox{1.0\columnwidth}{!}{      
\begin{tabular}{cccccccc}
\toprule
\multicolumn{3}{c}{IA-SSM}& \multirow{2}{*}{mAP}&\multicolumn{3}{c}{OA-SSM}&\multirow{2}{*}{mAP} \\
\cmidrule(r){1-3} \cmidrule(r){5-7}
 C3  & C4 & C5 &  & P3  & P4 & P5 &  \\ 
\midrule
  &   &   & 52.3&   &   &   &52.3\\
\checkmark &   &   & 53.8& \checkmark &   &   &53.9\\
  &\checkmark   &   & 54.0&   &\checkmark   &   & 53.9\\
  &  & \checkmark   & 54.3&   &  & \checkmark   & 54.1 \\
\bottomrule
\end{tabular} }
\vspace{-16pt}
\end{table}


\section{Conclusion}
In this paper, we propose DA-Mamba, a novel hybrid CNN-SSM architecture for DAOD. 
By integrating State Space Models (SSMs) with CNNs, DA-Mamba efficiently captures both local and global domain-invariant features.
Specifically, we incorporate Image-Aware SSM (IA-SSM) into the backbone for supplementing global domain attributes at the image-level, and Object-Aware SSM (OA-SSM) into the detection head to model spatial and semantic dependencies at the instance-level.
Experimental results show that DA-Mamba significantly improves cross-domain performance, outperforming existing methods while maintaining high computational efficiency.





\newpage

{
    \small
    \bibliographystyle{ieeenat_fullname}
    \bibliography{main}
}


\clearpage

\clearpage

\section{Appendix}

\subsection{Implementation Details}
We introduce the one-stage detector YOLO-World~\cite{YOLO-W} into the unsupervised domain adaptation (UDA) setting as the baseline detector.
In each iteration, one batch of source images with ground truth and one batch of target domain images with pseudo labels are forwarded to calculate the classification, adversarial and regression loss.
The hyperparameter $r, \lambda^I, \lambda^O$ is set to $2.0, 1.0, 0.5$.
The batch size for each domain is set to $2$, using the SGD optimizer with linear warm-up and decay learning rate.
The base learning rate is 1e-3, the warm-up period is 5 epochs and the decay starts from the 20th epoch.
Mean Average Precision (mAP) with a threshold of 0.5 is taken as the evaluation metric. 
All experiments are deployed on 1 Tesla V100 GPUs. 

\subsection{Additional Benchmarks}

\begin{table}[h]
\centering
\caption{Comparison ($\%$) on benchmarks with various categories.}
\label{tab-SF}
\setlength\tabcolsep{4pt}
\resizebox{1.0\columnwidth}{!}{      
\begin{tabular}{ccccc}
\toprule
Benchmark & Category & Baseline (UDA)& DA-Mamba & Gains \\
\midrule
C$\rightarrow$F   & 8  & 52.3 & 58.1 & 5.8 \\
C$\rightarrow$B   & 7  & 41.9 & 48.7 & 6.8 \\
P$\rightarrow$Clp & 20 & 46.2 & 52.5 & 6.3 \\ 
P$\rightarrow$Cmc & 6  & 37.9 & 43.8 & 5.9 \\
K$\rightarrow$C   & 1  & 60.4 & 62.5 & 3.1 \\
S$\rightarrow$C   & 1  & 59.3 & 62.3 & 3.0 \\  
\bottomrule
\end{tabular}}
\end{table}

\noindent \textbf{Evaluation on Multi-Category Benchmarks} In the main text, we report experiments on four mainstream multi-category adaptation scenarios.
As shown in Table~\ref{tab-SF} (rows $2\sim5$), DA-Mamba consistently achieves substantial improvements of $5.8\sim6.8\%$, demonstrating strong cross-domain generalization under diverse domain shifts.

\noindent \textbf{Evaluation on Single-Category Benchmarks} Beyond these benchmarks, KITTI~\cite{KITTI} and Sim10K~\cite{SIM10K} are also two important mainstream benchmarks. 
KITTI includes 7,481 real-world images with different camera Field of View (FoV) settings.
The synthetic dataset SIM10k has 10,000 photos from the GTA V video game, designed to evaluate synthetic-to-real transfer.
However, both adaptation scenarios KITTI$\rightarrow$Cityscapes (K$\rightarrow$C) and SIM10k$\rightarrow$Cityscapes (S$\rightarrow$C) contain a single category ``car''. 
These adaptation settings do not include inter-class semantic structures, and thus cannot fully evaluate DA-Mamba’s ability to model global semantic relationships across categories.
Consequently, compared with the multi-category scenarios (6–20 classes), DA-Mamba obtains smaller but still consistent improvements of $2.1\sim3.0\%$ on these single-category tasks. 
Importantly, since no cross-category semantic differences exist in these cases, the performance gain mainly comes from enhanced modeling of spatial dependencies under domain shift.
This result further validates that DA-Mamba remains effective even without cross-category semantics and can robustly model long-range spatial dependencies in single-category scenarios.

Overall, these results show that DA-Mamba provides consistent gains across both semantic-rich and -scarce adaptation settings, highlighting its robustness and broad applicability under diverse domain shifts.

\subsection{Structure Variants} 

\begin{figure}[h]
\centering
\includegraphics[width=1.0\columnwidth]{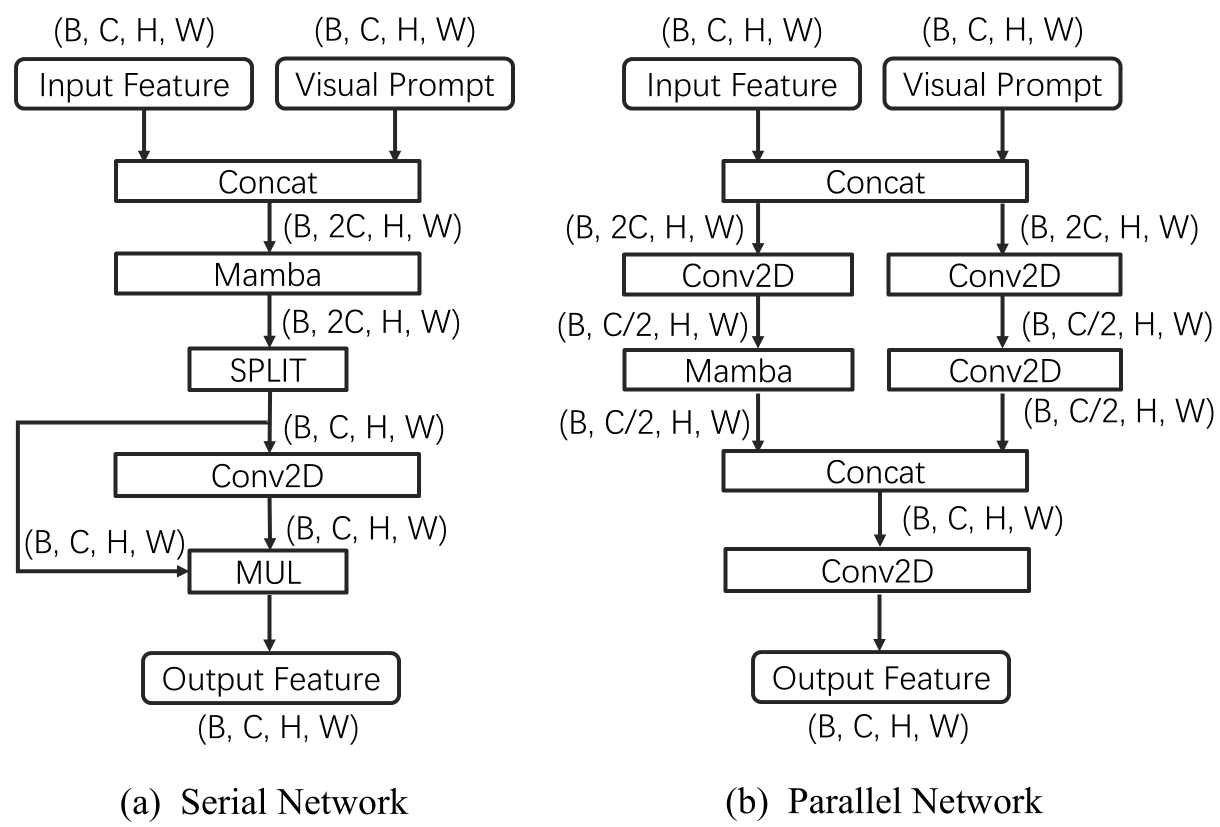} 
\vspace{-5pt}
\caption{Different structure for IA-SSM and OA-SSM.}
\label{fig4}
\vspace{-10pt}
\end{figure}

\begin{table}[h]
\centering
\caption{Comparison ($\%$) of different SSM structure.}
\label{tab-architecture}
\setlength\tabcolsep{14pt}
\resizebox{1.0\columnwidth}{!}{      
\begin{tabular}{cccc}
\toprule
Structure & Param(M)  & FPS & mAP \\  
\midrule
Serial   & 8.610 &  12.6   & 58.1 \\
Parallel & 1.732 &  14.1   & 58.1 \\
\bottomrule
\end{tabular}   }
\end{table}

\begin{figure*}[h!]
\centering
\includegraphics[width=1.0\textwidth]{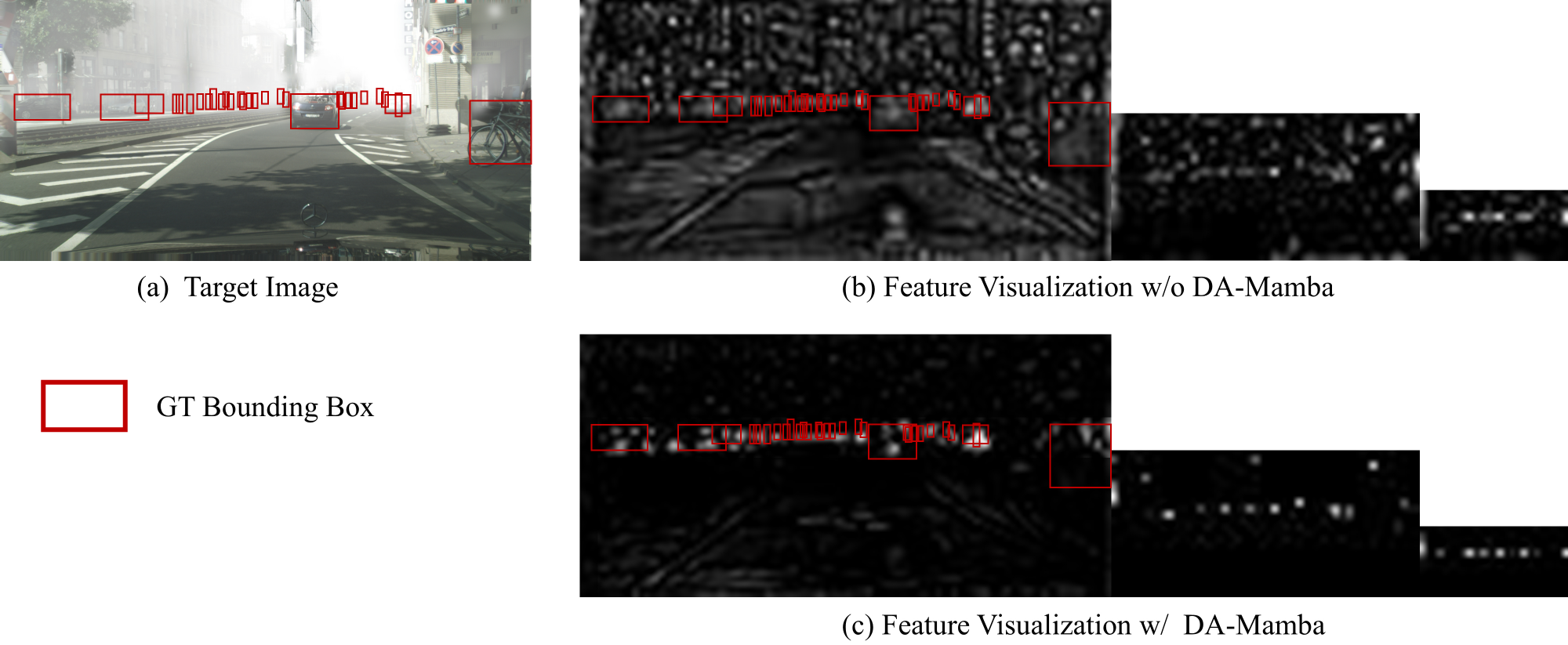} 
\caption{
Visualizations of the extracted feature map.
}
\label{fig-F-V}
\end{figure*}

\begin{figure*}[h!]
\centering
\includegraphics[width=0.7\textwidth]{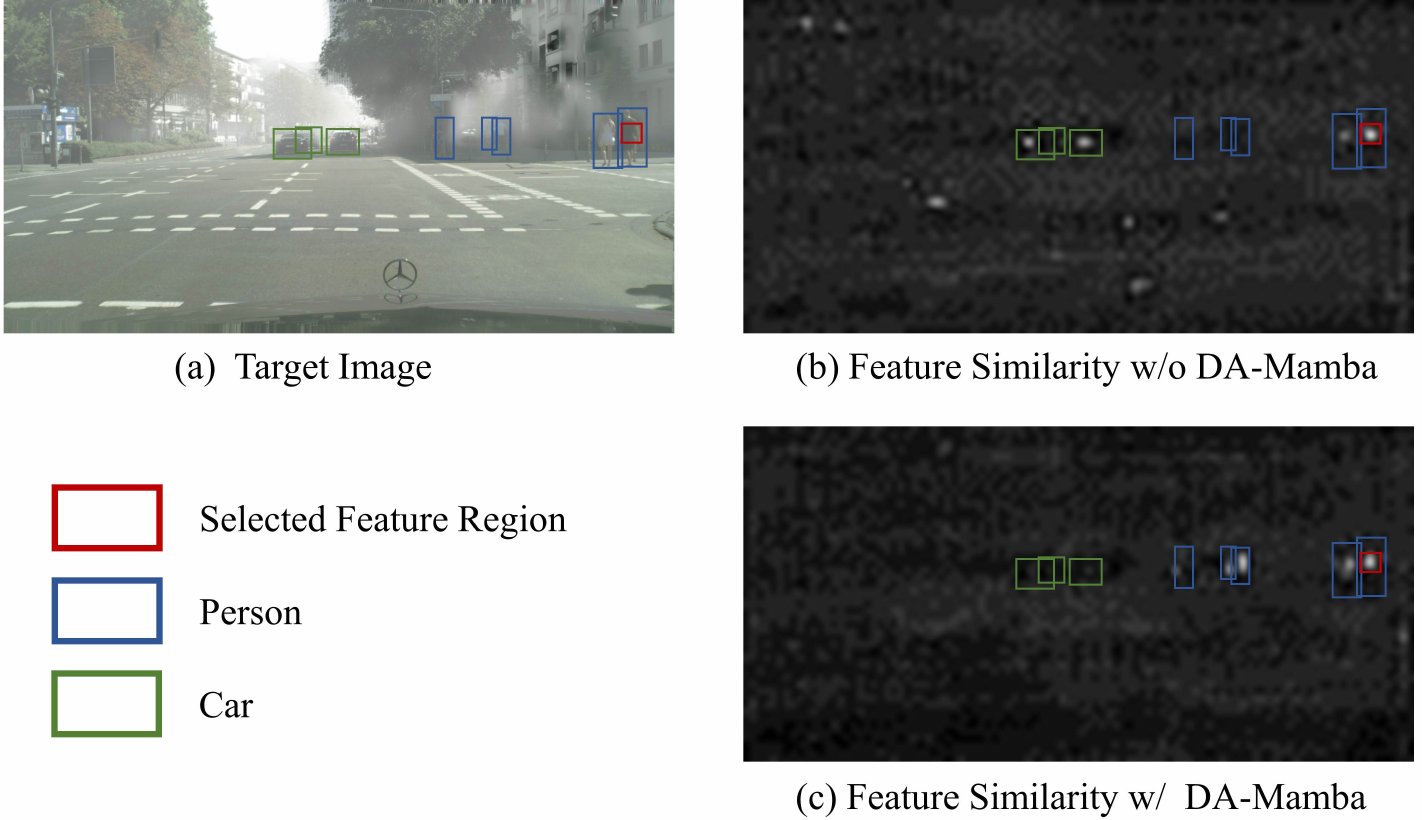} 
\caption{
Visualizations of the similarity between the features of a certain point and other regions on the feature map.}
\label{fig-F-S}
\end{figure*}

\noindent \textbf{Serial or Parallel?} \textit{Parallel network achieves comparable accuracy to the serial design with higher computational efficiency.}
Depending on the ordering of global and local alignment, we explore two network variants: (a) serial network and (b) parallel network.
As shown in Fig.~\ref{fig4}(a), the serial network first feeds the concatenated input-prompt features into the Mamba layer to learn global visual dependencies, and subsequently applies gated convolution to extract local domain-invariant features.
In contrast, the parallel network in Fig~\ref{fig4}(b) exploits a dual-pipeline design, where local and global information are learned independently through convolution pipeline and Mamba pipeline, and then fused to form the output feature.
As reported in Table~\ref{tab-architecture}, the parallel network requires only $20\%$(1.732M) of the parameters of the serial network(8.610M) while achieving the same mAP. 
It also reaches a higher inference speed (14.1 FPS vs. 12.6 FPS), demonstrating better computational efficiency.
This efficiency gain comes from avoiding the processing of expanded feature channels (2C) through both Mamba and convolution branches and instead keeping them lightweight (C/2) before fusion.

Given its superior efficiency and its natural alignment with the dual-pipeline modeling objective of DA-Mamba (\ie, global-local alignment), we adopt the parallel network structure in the design of IA-SSM and OA-SSM.

\begin{figure*}[h]
\centering
\includegraphics[width=0.85\textwidth]{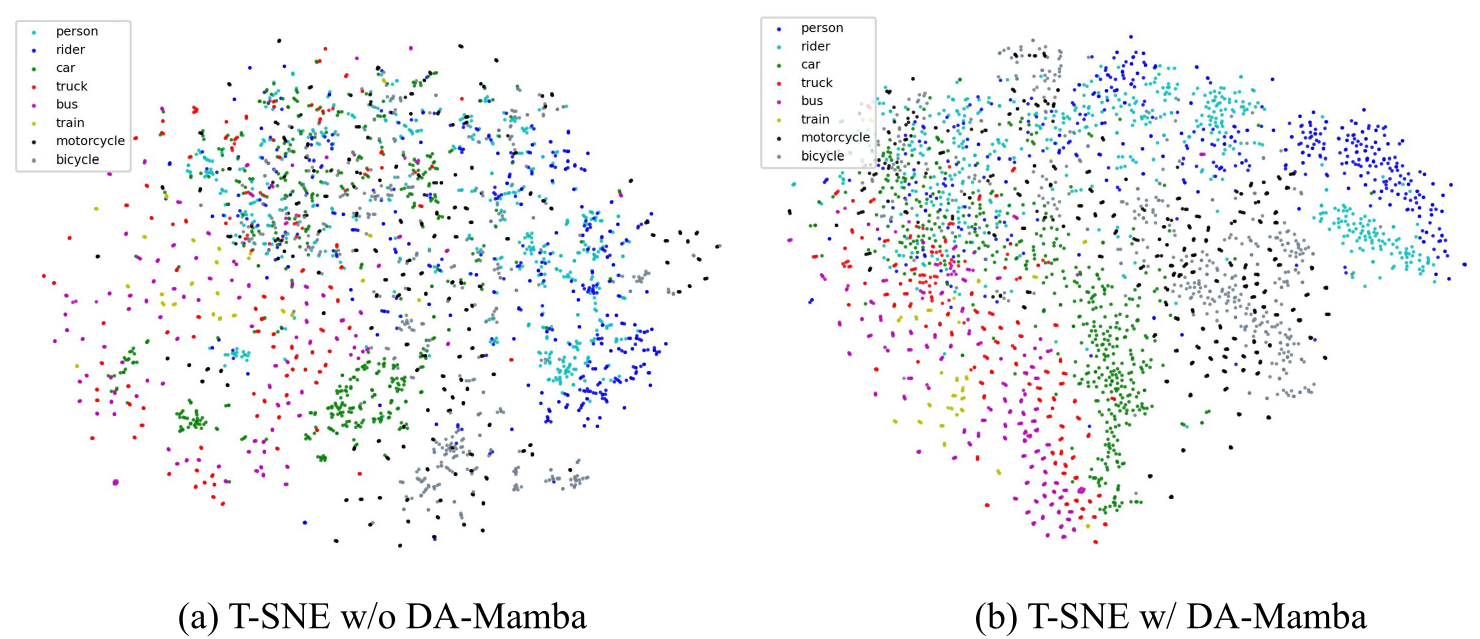} 
\caption{
T-SNE visualization of the features extracted from the target domain.
}
\label{fig-TSNE}
\end{figure*}

\subsection{Feature Visualization}

\textbf{Feature Visualization} \textit{DA-Mamba accurately extracts the features of foreground objects in the target domain while suppressing redundant, semantically-irrelevant background information.}
We visualize the multi-scale features before and after inserting the IA-SSM and OA-SSM modules.
As shown in Fig.~\ref{fig-F-V} (b), without DA-Mamba, the extracted features respond strongly to background regions while failing to highlight foreground objects, indicating insufficient domain alignment.
In contrast, Fig.~\ref{fig-F-V} (c) shows that DA-Mamba suppresses irrelevant textures (such as lane markings and background building) and emphasizes true foreground areas, suggesting more compact and domain-invariant representations.
This improved localization and background suppression reduce feature noise transferred across domains, thereby enhancing cross-domain generalization.

\noindent \textbf{Feature Similarity} \textit{DA-Mamba exhibits high intra-class similarity and low inter-class similarity across spatially distant regions, reflecting its capability in capturing global domain-invariant features and long-range semantic relationships.}
To intuitively verify DA-Mamba's ability to capture global domain-invariant features and capture long-distance semantic dependencies, we select a "Person" object from the target image (red boxes), and compute its similarity with all other regions in the feature map.
Fig.~\ref{fig-F-S}(b) and (c) show the visualization results of baseline and DA-Mamba, respectively, where brighter regions indicate higher similarity and darker regions indicate lower similarity.
As shown in Fig.~\ref{fig-F-S}(b), the baseline exhibits low similarity even within the same class (blue boxes) and undesired high similarity with other categories (green boxes), indicating weak global semantic discrimination.
Conversely, Fig.~\ref{fig-F-S}(c) shows that DA-Mamba consistently highlights same-class regions (blue boxes) while suppressing cross-category correlations (green boxes), regardless of spatial distance.
This demonstrates that the global-local alignment in DA-Mamba effectively propagates semantic dependencies over long ranges, showing a good ability to extract global domain-invariant representations that are robust to domain shifts.

Together, these qualitative results confirm that DA-Mamba achieves more stable and consistent domain alignment by simultaneously enhancing local domain-invariant features and global semantic-spatial dependencies, thereby providing robust cross-domain representations.

\subsection{T-SNE Visualization}

\begin{figure*}[h]
\centering
\includegraphics[width=0.95\textwidth]{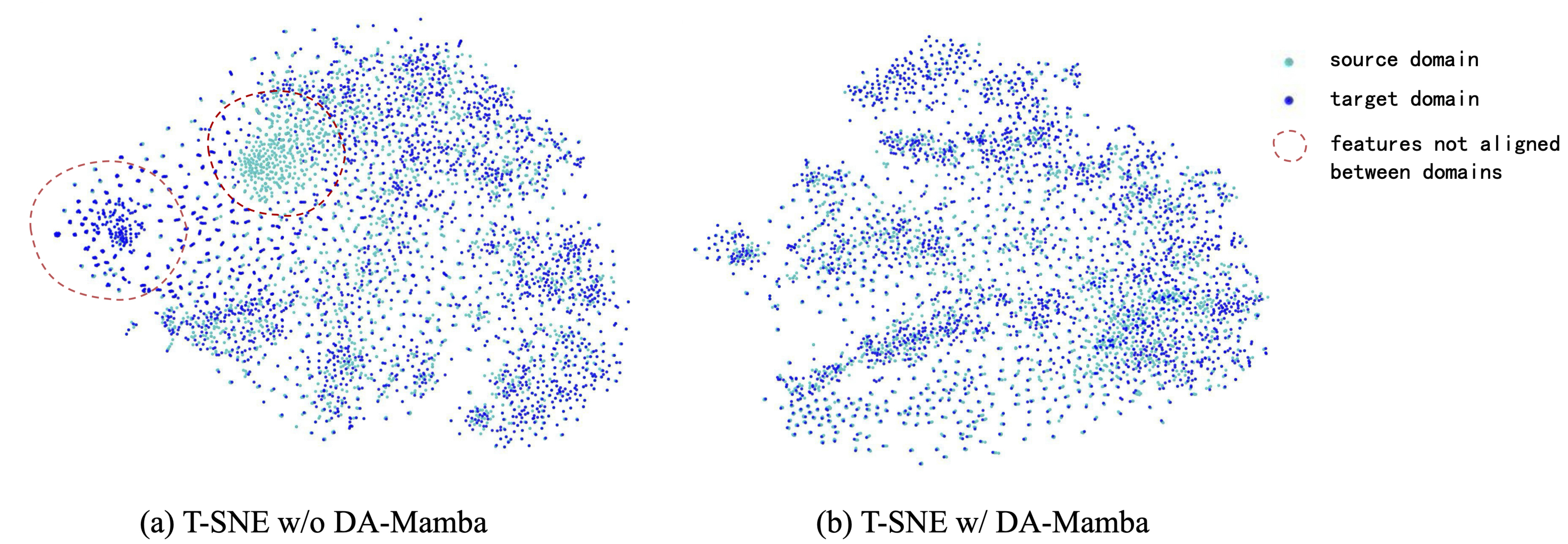} 
\caption{
T-SNE visualization of the features extracted from both domains.}
\label{fig-TSNE-ST}
\end{figure*}
\noindent\textbf{T-SNE on the Target Domain.} 
\textit{The features extracted by DA-Mamba exhibit tighter intra-class clusters and larger inter-class margins, reflecting better semantic structuring in the target domain.}
To intuitively demonstrate the effect of DA-Mamba, we provide a T-SNE visualization of features extracted from the target domain.
As shown in Fig.~\ref{fig-TSNE}(a), before introducing DA-Mamba, features from different categories are heavily mixed and instances of the same category do not form compact clusters.
In contrast, Fig.~\ref{fig-TSNE}(b) shows that DA-Mamba produces clear boundaries between categories.
At the same time, categories with strong semantic relevance lie closer in the embedding space, such as person–rider, car–truck, and bicycle–motorcycle.
Moreover, categories that are spatially related in the scene, such as rider, bicycle, and motorcycle, are also grouped nearby.
This indicates that benefiting from global–local alignment, DA-Mamba effectively learns both semantic and spatial dependencies, leading to more structured and discriminative representations in the target domain. 

An interesting observation is that the rider category is split into two sub-clusters, which we attribute to the semantic difference between riders on bicycles and riders on motorcycles.

Finally, we note that both (a) and (b) contain a region where different categories are mixed, mainly due to heavy occlusions and visually inseparable objects in those cases.

\noindent\textbf{T-SNE across Source and Target Domains.} 
\textit{DA-Mamba achieves better feature alignment between the source and target domains.}
We further visualize the T-SNE of features from both domains in Fig.~\ref{fig-TSNE-ST}.
As highlighted by the red dashed circles in Fig.~\ref{fig-TSNE-ST}(a), features from the source domain (cyan) are not well aligned with those from the target domain (blue), indicating that the baseline model fails to achieve sufficient cross-domain alignment.
In contrast, with the proposed DA-Mamba, features from both domains in Fig.~\ref{fig-TSNE-ST}(b) are tightly mixed and share similar distributions.
This enhanced alignment results from DA-Mamba's global–local alignment mechanism, where global SSM pipeline establishes domain-invariant long-range structure while local convolutional pipeline anchors the alignment to stable low-level patterns.
Such complementary alignment enables the features of both domains to be aligned into a shared, well-organized distribution.

Overall, these T-SNE visualizations demonstrate that DA-Mamba fundamentally improves cross-domain feature alignment, that is, better semantic structure in the target domain and better alignment across source and target distributions, confirming that its performance gains primarily arise from fine-grained global–local domain-invariant representation learning rather than merely improving the detector capacity.

\subsection{Detection Results}
\begin{figure*}[h]
\centering
\includegraphics[width=1.0\textwidth]{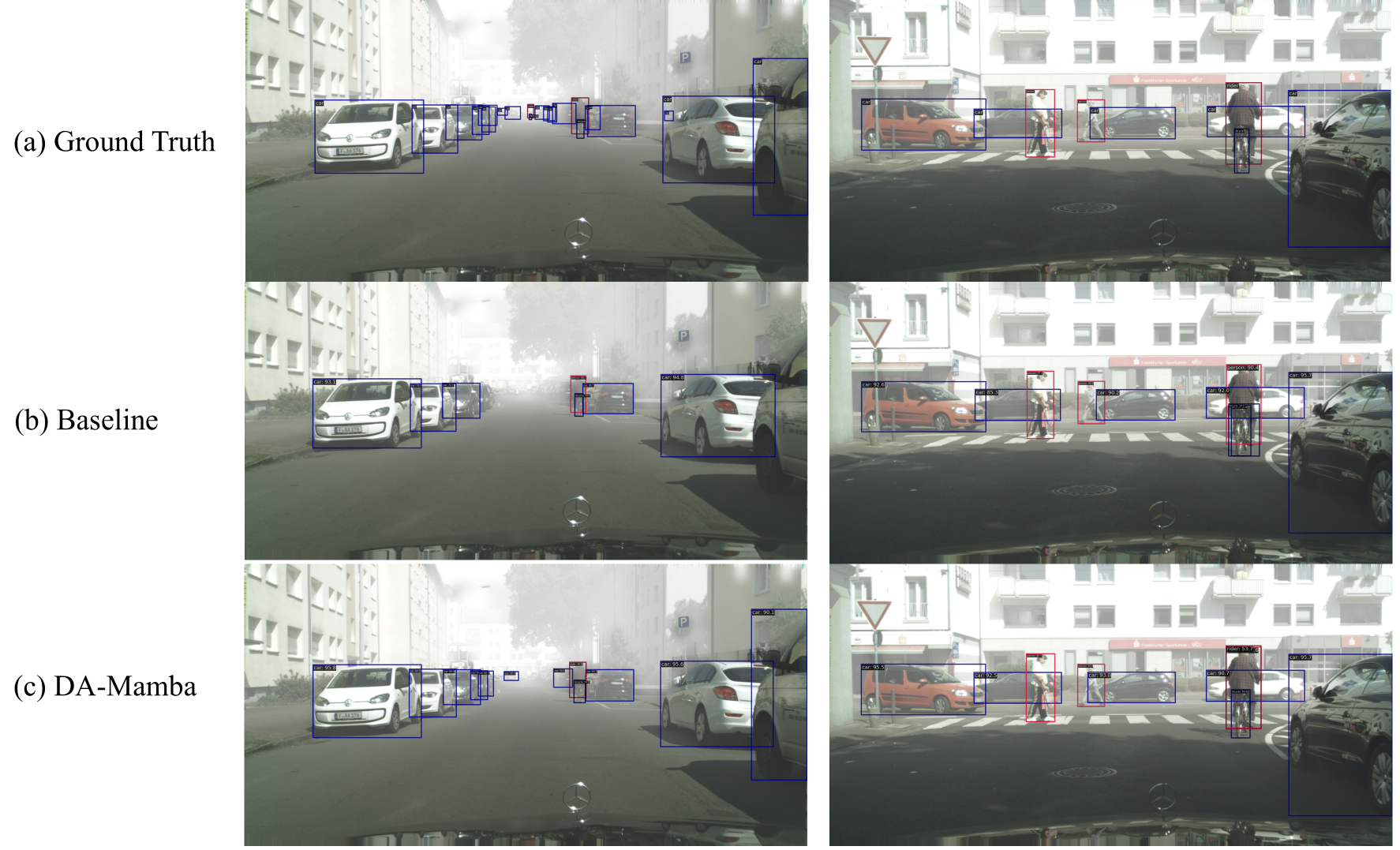} 
\caption{
Detection results visualization.
}
\label{fig-Detection}
\end{figure*}
We visualize the detection results of baseline and DA-Mamba in Fig.~\ref{fig-Detection}
While baseline misclassifies the rider as person, our DA-Mamba correctly classifies the rider.
Meanwhile, DA-Mamba shows higher prediction confidence, and detects more objects in the fog that are missed in the baseline.
This indicates that DA-Mamba effectively learns local and global domain properties, achieving fine-grained domain alignment.

\subsection{Pseudo Label}

\begin{table}[h]
\centering
\caption{Analysis ($\%$) on Pseudo Labels.}
\label{tab-Loss}
\setlength\tabcolsep{4pt}
\resizebox{1.0\columnwidth}{!}{      
\begin{tabular}{cccccc}
\toprule
Benchmark & Pseudo Labels & C$\rightarrow$F &  C$\rightarrow$B   & P$\rightarrow$Clp & P$\rightarrow$Cmc\\
\midrule
Baseline(UDA) &              & 50.6 & 40.0 & 43.8 & 35.9\\
Baseline(UDA) & $\checkmark$ & 52.3 & 41.9 & 46.2 & 37.9\\
DA-Mamba &              & 57.1 & 48.0 & 51.0 & 42.5\\
DA-Mamba & $\checkmark$ & 58.1 & 48.7 & 52.5 & 43.8\\
\bottomrule
\end{tabular}}
\end{table}

\noindent\textbf{Effect of Pseudo Labels.}
\textit{DA-Mamba does not rely on pseudo labels for performance gains.}
Table~\ref{tab-Loss} reports the results of applying pseudo label in the target domain (\ie, the loss term $\mathcal{L}_{cls}^T$) during training.
For the baseline UDA detector, introducing pseudo labels yields noticeable improvements ($1.7\sim2.4\%$), reflecting its strong dependence on target domain supervision to compensate for insufficient domain alignment.
In contrast, DA-Mamba already achieves significant improvements without pseudo labels, and adding pseudo labels provides only marginal additional gains ($0.7\sim1.5\%$).
This smaller improvement margin indicates that DA-Mamba substantially reduces the domain discrepancy itself, leaving less room for pseudo labels to correct target domain errors.
Therefore, the performance gains of DA-Mamba do not stem from pseudo labels but from its stronger global–local domain alignment capability.

\subsection{Failure Case}
We observe that the missed detections of DA-Mamba mainly occur in regions with strong ambiguity or heavy occlusion, as shown in the middle area of Fig.~\ref{fig-Detection}. 
Since such cases suffer from severely degraded visual semantics, similar failures are commonly observed in most object detection frameworks and are not specific to our method.

\subsection{Error Bars}
\begin{table}[h]
\centering
\caption{ mAP($\%$) on four benchmarks.}
\label{Error_bar}
\setlength\tabcolsep{10pt}
\resizebox{1.0\columnwidth}{!}{      
\begin{tabular}{ccccc}
\toprule
C$\rightarrow$F & C$\rightarrow$B & P$\rightarrow$Clp & P$\rightarrow$Cmc\\
\midrule
58.1($\pm$ 0.2) & 48.7($\pm$ 0.4) & 52.5($\pm$ 0.3) & 43.8($\pm$ 0.4) \\
\bottomrule
\end{tabular}  }
\end{table}
We provide error bars in Table~\ref{Error_bar}. 
The error bars are captured by multiple running with given experimental conditions.

\subsection{Extended Discussion on Global-Local Alignment Design}

\noindent\textbf{Why DA-Mamba Outperforms Transformer-Based DAOD methods and Convolutional Attention Variants.}
Table~\ref{tab-computational} in the main text shows that DA-Mamba not only surpasses the SOTA transformer-based DAOD method DATR~\cite{DATR} with significantly lower computational overhead, but also outperforms convolutional attention variants under similar cost.
This advantage stems from DA-Mamba’s novel synergistic integration of convolutional locality and SSM-based global perception, enabling efficient and fine-grained global–local alignment.

Transformer-based DAOD methods introduce self-attention to capture global dependencies across domains. 
However, their quadratic complexity forces the use of large patch sizes to reduce token counts.
Since the same patch is treated as a whole in attention computation, larger patches make the granularity of attention computation coarser.
Moreover, Transformer attention is domain-agnostic, treating all positions uniformly and failing to emphasize locally stable domain-invariant features, which are crucial for local alignment.
In contrast, DA-Mamba employs a dedicated convolutional pipeline that preserves fine-grained spatial detail and efficiently extracts local domain-invariant features.

Convolutional attention variants attempt to approximate global interactions with lower cost but rely on globally shared parameters.
For example, GCNet~\cite{GCNet} discards Q(query) and only use K(key) and V(value) to simplify attention calculation.
However, eliminating the query branch prevents pixel-level semantic analysis and makes the model sensitive to domain bias.  
And HAM~\cite{HAM} decomposes attention into spatial attention and channel attention based on convolution and pooling, which is restricted by the sliding window receptive fields and inherently cannot capture semantic dependencies between distant objects.  
Such shared global descriptors suppress domain-specific variations and are easily biased by style differences between source and target domains, causing insufficient global alignment. 
In contrast, the SSM pipeline in DA-Mamba performs a directed scan across the entire feature map, allowing each position to aggregate information from all previous positions.  
This gives DA-Mamba the ability to model long-range semantic and spatial dependencies across domains.

Overall, DA-Mamba adopts a global–local hybrid design: 
(1)~the convolution pipeline efficiently extracts stable local domain-invariant features through local connectivity and translation invariance, and  
(2)~the SSM pipeline provides linear-time global modeling of long-range dependencies.
This decomposition allows DA-Mamba to decouple the domain adaptation process into local alignment and global alignment, avoiding both the domain-agnostic behavior of Transformers and the rigidity of convolutional attention, leading to more effective cross-domain representation learning.

\noindent\textbf{Why Enlarging the Receptive Field Promotes Domain Alignment.}
Domain adaptive object detection typically optimizes an adversarial domain classifier using an MSE loss:
\begin{equation}
    \label{MSE}
    \mathcal{L}_{\text{adv}} = 
    \frac{1}{HW} \sum_{i=1}^{H}\sum_{j=1}^{W} 
    \left(\mathcal{D}(\mathbf{f}_{ij}) - y\right)^2,
\end{equation}
where $\mathbf{f}_{ij}$ denotes the feature at location $(i,j)$ on the feature map $\mathbf{f}$, the domain classifer $D(\cdot)$ predicts the probability that a feature comes from the source domain, and $y\in\{0,1\}$ is the domain label.  
This objective enforces the feature distributions of the two domains to be indistinguishable at \emph{every} pixel position.

When the receptive field is small, $\mathbf{f}_{ij}$ contains only local information, \ie the content around $(i,j)$ on feature map $\mathbf{f}$.  
Formally, if $\mathbf{f}_{ij}^d$ depends only on a local region $\mathbf{N}^d_{ij}$, we have
\begin{equation}
    \mathbf{f}^d_{ij}=\mathcal{F}(\mathbf{N}^d_{ij}),
\end{equation}
where $d\in\{S, T\}$ denotes the source or target domain, and $\mathcal{F}$ denotes the visual encoder.
In this case, the domain classifier processes each spatial location independently, without explicitly modeling relationships between different regions.
Therefore, the adversarial loss in Eq.~\eqref{MSE} encourages
\begin{equation}
    \label{P1}
    p\big(\mathcal{F}(\mathbf{N}^S_{i_s j_s})\big) = p\big(\mathcal{F}(\mathbf{N}^T_{i_t j_t})\big),
\end{equation}
where $p(\cdot)$ denotes the feature distribution.
However, Eq.~\eqref{P1} only enforces domain invariance of \emph{local} features within each region, and does not constrain the global relationships among regions.
Therefore, such alignment remains local and is insufficient to capture global domain shifts (\eg, object co-occurrence, layout, or scene style).

When the receptive field covers the entire image, each $\mathbf{f}_{ij}^d$ additionally aggregates information from a broader semantic context:
\begin{equation}
\mathbf{f}^d_{ij} = 
\big[\mathcal{F}(\mathbf{N}^d_{ij});\,\mathbf{c}^d_{ij}\big],
\end{equation}
where 
\begin{equation}
\mathbf{c}^d_{ij} = \mathcal{S}\Big(\big\{\mathbf{N}^d_{kl}\big\}_{k=1,l=1}^{H,W}\setminus\{\mathbf{N}^d_{ij}\}\Big)
\end{equation}
contains features from all other regions via a global modeling network $\mathcal{S}$.
In this setting, each $\mathbf{f}^d_{ij}$ simultaneously encodes its own local feature and its relationships to all other regions, including object structure, inter-object relations, and scene layout.
After optimizing Eq.~\eqref{MSE}, we then expect
\begin{equation}
\begin{split}
    \label{P2}
    p\big(\mathcal{F}(\mathbf{N}^S_{i_s j_s})\big) &= p\big(\mathcal{F}(\mathbf{N}^T_{i_t j_t})\big),\\
    p\big(\mathbf{c}^S_{i_s j_s}\big) &= p\big(\mathbf{c}^T_{i_t j_t}\big).
\end{split}
\end{equation}
This means that, for each local region, both its own local features and its global relationships with other regions are required to be domain-invariant, thereby achieving \emph{simultaneous local and global alignment}.

In summary, enlarging the receptive field allows the adversarial loss to act on both local semantics and long-range dependencies at each spatial location, effectively promoting alignment between the two domains.
In DA-Mamba, this principle is instantiated by the dual-pipeline design: the convolution pipeline corresponds to $\mathcal{F}(\mathbf{N}^d_{ij})$ and focuses on local domain-invariant information, while the SSM pipeline corresponds to $\mathbf{c}^d_{ij}$ and models global semantic and spatial dependencies.
Their fusion enables DA-Mamba to implement global–local domain alignment in a theoretically grounded manner, which explains its superior cross-domain performance in our experiments.

\noindent\textbf{Future Work.}
While DA-Mamba demonstrates that global–local alignment can significantly improve cross-domain generalization, several directions remain promising for future exploration.
First, instead of inserting SSM blocks into FPN as in the current design, enabling SSM with native multi-scale global modeling capability may further strengthen global semantic propagation across resolutions.
Second, integrating uncertainty estimation schemes could allow the model to dynamically balance global and local alignment according to the degree of domain shift.
Finally, applying the proposed global–local alignment principle to other adaptation tasks such as instance segmentation, open-vocabulary detection, or video-domain adaptation may broaden the applicability of DA-Mamba.
We leave these directions for future investigation.

\end{document}